\documentclass{article}    

\usepackage{authblk}
\usepackage{graphicx}
\usepackage{amsmath}
\usepackage{amssymb}
\usepackage{algorithm}
\usepackage{algpseudocode}
\usepackage{scalerel}
\usepackage{xurl}
\usepackage{multicol}
\usepackage{dcolumn}
\usepackage{float}
\usepackage{booktabs}
\usepackage{caption}
\usepackage{subcaption}
\usepackage[linguistics]{forest}
\usepackage[misc,geometry]{ifsym}
\usepackage[round]{natbib}

\begin{document}

\title{Measure Inducing Classification and Regression Trees for Functional Data}

\author{Edoardo Belli \\
        edoardo.belli@polimi.it \and Simone Vantini \\ 
                                     simone.vantini@polimi.it}

\affil{MOX - Modeling and Scientific Computing, Department of Mathematics, Politecnico di Milano, Italy}

\date{}

\maketitle

\begin{abstract}
We propose a tree-based algorithm for classification and regression problems in the context of functional data analysis, which allows to leverage representation learning and multiple splitting rules at the node level, reducing generalization error while retaining the interpretability of a tree. This is achieved by learning a weighted functional $L^{2}$ space by means of constrained convex optimization, which is then used to extract multiple weighted integral features from the input functions, in order to determine the binary split for each internal node of the tree. The approach is designed to manage multiple functional inputs and/or outputs, by defining suitable splitting rules and loss functions that can depend on the specific problem and can also be combined with scalar and categorical data, as the tree is grown with the original greedy CART algorithm. We focus on the case of scalar-valued functional inputs defined on unidimensional domains and illustrate the effectiveness of our method in both classification and regression tasks, through a simulation study and four real world applications.
\end{abstract}

\section{Introduction}
\label{intro}
The idea of making predictions using hierarchical, binary if-else decision rules is naturally appealing to rational human agents, as it mimics our way of thinking on a simplified and abstract level. This is also reflected in the way we program our machines, where most programming languages make use of some type of conditional branching. In the artificial intelligence community, human defined rules where used to build expert systems that where able to leverage large knowledge bases, with the objective to aid or even replace decision makers. With the increase in computational power, memory capacity and data collection, it became clear that the approach of learning those rules was far superior than designing them. A major contribution in this direction, was the introduction of decision tree learning, and in particular the classification and regression trees (CART) algorithm \citep{book_cart}, which sprouted a whole new literature on tree-based methods in machine learning. The success of the CART framework was mainly driven by its flexibility and interpretability, as it provided a nonparametric tool with nonlinear decision boundaries for both classification and regression problems, which was also able to handle categorical and numerical variables, while at at the same time retaining the intuition of binary splitting rules. From the generalization perspective, CART always suffered from instability issues, where small variations in the input data could lead to inconsistent predictions. Successful techniques like bagging \citep{bagging} random forest (RF) \citep{randomforest} and gradient boosted trees (GBT) \citep{boosting} were introduced to reduce variance and bias, improving performance at the expenses of interpretability. \\
In the last decade, data has become more complex and new tools are needed to extract meaningful and reliable information from it. When sample sizes are large and the focus is on prediction, deep learning (DL) has proven to be the best approach, with an overwhelming success in multiple applications that deal with complex data types, like computer vision and natural language processing. Thanks to their scalable and modular architectures, DL methods are able to exploit huge sample sizes to learn representations that capture the inner structure of these data and generalize well in different settings. However, when data is scarce and interpretability is sought after, DL is not an option and we need an additional modeling effort to be able to handle complex data. Functional Data Analysis (FDA) can be seen as a data model for smooth and high dimensional data, which often arises in applications where high-frequency sampling devices are involved, like in electrical engineering, chemometrics and in the biomedical field. The typical instance of functional data is a scalar quantity measured in time, but in general we are dealing with multidimensional objects defined on multidimensional domains, like RGB images or volumes. From a practical standpoint, functional data is clearly discrete and finite dimensional, as it is collected by sensors and stored in memory. This however should not preclude the opportunity to take advantage of the continuous nature of the phenomena that generate these data, like in the case of physical quantities. A naive approach is to look at the raw data as p evaluations of a function on a discretized grid over the domain, which does not take into account the fact that different samples could have been generated by grids with different spacing, even if they have the same number
of points p. Also, there could be multiple missing values and each evaluation could be affected by noise induced by the sensors used for collecting the data, including small time discrepancies between different readings. Therefore, the true underlying functions are expressed as a finite basis expansion and are usually recovered by means of smoothing or functional principal components, depending on the application and on the amount of missing values. Another common practice is to align the input functions to further reduce the phase variability and retain only the information given by the amplitude variability \citep{fda_kma}. Once the basis and/or the coefficients are estimated in this preprocessing step, the functions are evaluated on the same dense grid and known multivariate methods are employed for the numerical estimation of a classifier/regressor, with the limitations imposed by the intrinsic high dimensionality, which are not only related to model fitting and generalization error. In the case of CART, it is well known that in high dimensions the resulting trees can be prohibitively complex for any intuitive and significant interpretation, not only for the excessive height of the trees, as in some cases an intricate prediction rule may be considered more plausible \citep{ruleplausibility}, but especially for the difficulty of assigning a clear meaning to each split, defeating one of the main advantages of the method. Even in the case of axis-parallel splits, if for instance we consider a classification tree that has been trained on dense spectrometric input data, it may be contrived to justify selecting a specific wavelength as the splitting variable, while at the same time binning the domain in arbitrary subintervals may be too simplistic. In the FDA literature, much attention has been devoted to linear models, dimensionality reduction and inference \citep{book_fda,book_inferfda}. Regarding supervised learning, nonparametric methods with functional covariates have also been studied extensively \citep{book_nonparafda}, often in conjunction with other approaches like SVMs \citep{fda_svm}, domain segmentation \citep{fda_segm} and distance learning \citep{fda_weighted,fda_maha}. In the case of tree-based methods, most works focus either on functional inputs or responses, as the two aspects are in general decoupled, given that the response is related to the specific loss function for the task. The problem of dealing with functional covariates is often tackled by splitting each node with respect to the distance from a chosen template function, and we will discuss this more in detail in the next section. As data belongs to an infinite dimensional functional Hilbert space, rather than treating it as multiple evaluations over a dense grid, another approach is to consider each function as a whole unit \citep{fda_features}. This opens up the possibility of extracting global information from the functions and their derivatives, like integral features, without relying exclusively on the pointwise evaluations, which are high dimensional and could be affected by measurement error or extreme fluctuations of the underlying phenomena, as the true functions are never available but are always recovered through some estimation process. Following this approach, we focus on problems with with functional inputs and introduce a tree-based algorithm that extends the standard CART algorithm to deal with (multiple) functional covariates, with the objective of improving both prediction and interpretability, while at the same time keeping all the advantages of decision trees, like the option of including other categorical/scalar covariates, different pruning schemes and the choice of any loss function. This is achieved by learning multiple weight functions for each internal node, inducing a set of measures that are used to extract multiple integral features from the input data, effectively learning representations that are weighted $L^{2}$ spaces, as we will elaborate in section three. Finally, in section four we will analyze the performance in classification and regression problems with both simulations and real world applications.

\section{Related Work}
\label{related}
Starting from the early works that appeared more than three decades ago \citep{book_cart,dtinduction}, the literature about classification and regression trees has kept growing and these methods are still relevant nowadays, with developments that encompass multiple elements of the framework. In this section, we will focus on the three main aspects that we think are at the core of tree-based learning, in order to highlight the differences with respect to our proposal. It is clear that while presented separately, all three aspects are related to each other in many different ways, and we do not aim to provide an exhaustive analysis of the subject.
\subsection{Training}
The first aspect is the training process, which is related to the size and the architecture of the tree, that has to be controlled through hyperparameter optimization and/or model search. The original CART uses a greedy algorithm that is based on locally optimal binary partitions of the input space. This allows to avoid fixing a priori the height of the tree, which is restricted by means of pruning, either while growing the tree (pre-pruning) or after the tree has been fully grown (post-pruning). Empirical evidence shows that pre-pruning often leads to simpler trees with the same accuracy as post-pruned ones \citep{pmetrictreestopping}, where a typical approach is to control the growth of the tree by either limiting the maximum height or by fixing the minimum number of elements in a leaf. Permutation tests have also been used as an unbiased stopping criteria \citep{unbrecurpart}, with a p-value threshold as the hyperparameter to tune. If instead we accept to fix the height of the tree a priori, alternative approaches have been proposed in which the tree is trained for global optimality with different algorithms, like expectation maximization \citep{hme}, stochastic gradient descent with backpropagation \citep{effnongreedy,deepneuraldecforest} or by solving a mixed integer optimization problem \citep{optimaldt}. In the case of our proposed method, the tree is trained with the traditional greedy algorithm, which also depends on the weight functions that are learned for each internal node. Regarding this aspect, our method can be seen as a generalization of the original CART, as it does not prevent the usage of the univariate and multivariate splitting rules that are known in the literature, nor it forces the employment of any specific stopping or pruning scheme, while adding the option to deal with functional data at the node level without relying exclusively on the single evaluations.
\subsection{Node Splitting}
The second and most discussed aspect is how to perform the split inside a single node, which comprises both the problems of selecting a suitable splitting variable and computing the split itself. While CART is mostly known for axis parallel splits, which selects a single variable and therefore limits the shape of the decision boundary that can be learned, the original algorithm already allowed for oblique splits (CART-LC), by splitting with respect to a linear combination of the input features that was computed with an heuristic based on deterministic coordinate search. In the following years, multiple approaches have been proposed to find good and possibly sparse oblique split, including recursive least squares with feature elimination \citep{multivdt}, stochastic global optimization \citep{sadt,oc1,obtreeevo} and GLMs \citep{functionaltrees}. Other works focused exclusively on classification, inducing the splitting rule by the use of linear programming and SVMs \citep{disclinprog,enlarging,margintrees}, logistic regression \citep{logisticmodeltrees}, budget-aware classifier \citep{dtultrahigh} and different types of discriminant analysis \citep{discritreesclass,discritrees}. A known issue of CART is the one of selection bias, where variables with many possible splitting values are preferred during the selection process, favouring categorical variables and variables with fewer missing values. Therefore, the problem of unbiased split variable selection has been tackled by means of statistical hypothesis testing for both classification \citep{quest,cruise,ctreemissingvalues,guide_class} and regression \citep{guide_reg} trees, where the splitting variable with the lowest p-value is chosen, and then the split can be computed with any given approach. In the context of time series and FDA, most of the attention has been devoted to classification with one functional input, where for each internal node of the tree, one or more template functions are chosen and then a univariate split is performed with respect to the distance from such functions. The selection of the template functions has been implemented using exhaustive search \citep{dttimesstandard,dttimes}, binary clustering \citep{dttimesstandard,fda_dtfunctional} or by taking the mean function by class \citep{fda_classtfunctional}. This template technique has also been used together with recursive binary domain segmentation and multiple distance metrics \citep{dttimes}. The domain segmentation approach has further been used in a RF fashion for classification and regression with one functional input \citep{fda_rf}, where each tree of the forest is a standard CART that is fitted on a different preprocessed bootstrap subset of the inputs. This preprocessing step consists in randomizing a set of possibly overlapping intervals over the domain of the functions and taking the mean of the functions over these intervals, which is not related to the fact that the classifiers are trees and could be used in conjunction with other ensemble methods. We propose a node splitting procedure that can be used for both classification and regression, which is based on locally learning a weighted $L^{2}$ space by means of constrained convex optimization, that allows to extract multiple linear and nonlinear weighted integral features from the input functions. The concept of node-wise representation learning is not new, as there are works in computer vision that train a multilayer perceptron for each node of the tree \citep{neuraldecforest,deepneuraldecforest}. While a deep neural network learns a hierarchical and nonlinear representation of the input data, our representations are linear and not hierarchical and we need to specify the features that we want to extract for each node. For this reason, this can be seen as a model-based way of learning representations, which may lack the expressive power of more abstract ones, but has the advantage of retaining the interpretability, since the learned weight function is a measure of importance of the subsets of the domain of the input functions, and the extracted features can be nonlinear and have a clear mathematical meaning. Moreover, while the focus of this work is on the splitting rules for functional data, it is worth to note that as our rule fits in the original CART framework, in principle one could implement our approach to deal with the functional covariates, mapping each function to a real value, while at the same time including any other data source like categorical variables. This would introduce again the problem of biased variable selection, which in turn can be solved by resorting to any of the selection rules based on statistical hypothesis testing.
\subsection{Input and Response Data}
The third aspect is the type of task that the tree can solve and therefore the nature of the input and response data. The original CART is suited for both classification and regression problems with multivariate and categorical input data, and it has been extended to multivariate regression \citep{multivrespo}, longitudinal responses \citep{guidemultivrespo,boostdtlongdata} and other regression methods in the leaves \citep{regreleafs}, multitask learning \citep{multitask,ctreemultibinres,ctreemultilabel,ctreemultilabel2} with hierarchical labels \citep{dthierarchical}, high label dimensionality \citep{dtultrahigh} and responses in the form of class probabilities \citep{kdeleafs}. In the case of a single response that is a probability density function, the purity of the split has been evaluated by computing the sum of the f-divergences between the responses that correspond to the inputs in the left and right nodes \citep{fda_treeresponse_ocean}, while smoothing and principal component analysis have been used to reduce the dimensionality of the response as a preprocessing step \citep{fda_treeresponse_timeofday}. Tree-based methods have also been employed for conditional quantile estimation \citep{quantregtree,quantregforest} and conditional density estimation \citep{cdetree}. As previously noted, the objective of this work is to introduce a splitting approach for functional inputs in the context of the original CART algorithm, without specific attention to the type of response or any additional categorical and/or scalar data. In particular, we focus on the case of (multiple) scalar-valued functional covariates defined on a one-dimensional domain, but in principle our approach could be extended and tailored to specific applications with more complex data, like vector-valued functions defined on multidimensional domains.

\section{The Measure Inducing CART}
\label{mucart}
In this section, we describe our method with respect to functional data, but in general it can be applied to any Hilbert data, like in the usual multivariate case. The way our tree is trained is by recursive binary splitting of the input space, as in the original CART. For this reason, we will focus on the node splitting procedure. Regarding pruning, we control the growth of the tree by limiting the minimum number of elements in a leaf, without any post pruning technique like cost complexity pruning and weakest link pruning. This is just an implementation choice and we are not restricted to the use of any patricular pruning technique. In order to underline the differences with the functional case and establish notation, we briefly recall the multivariate CART algorithm for splitting a node.\\

Let $\mathcal{D}=\lbrace (x_i,y_i) \rbrace_{i=1}^{N}$ be the training set with $x_i \in \mathbb{R}^p$ and $y_i \in \{1,..,K\}$ for classification or $y_i \in \mathbb{R}$ for regression. Let $j=1,..,p$ and $s\in \mathbb{R}$ a splitting threshold, split the input space into two subregions such that

\begin{center}
$\mathcal{R}_1(j,s) = \lbrace  x | x_j \leq s \rbrace   \:\:\:\:\:\:\:\:\:   \mathcal{R}_2(j,s) = \lbrace  x | x_j > s \rbrace$
\end{center}

where $j$ and $s$ are chosen by solving the following optimization problem:

\begin{equation}
\label{eq:cart}
\underset{j,s}{\Huge{min}} \: \Big{[} \mathcal{L}\left(\mathcal{R}_1(j,s)\right) + \mathcal{L}\left(\mathcal{R}_2(j,s)\right) \Big{]}
\end{equation}

and $\mathcal{L}$ is the loss function of the node which depends on the task, for example gini index for classification and mean-square error for regression. To compute the optimal regions $\mathcal{R}_{1}(j^{*},s^{*})$ and $\mathcal{R}_{2}(j^{*},s^{*})$, we can find $j^{*}$ and $s^{*}$ by enumerating through the features for $j=1,..,p$ and through the the splits $s_{i}=x_{ij}$ for $i=1,..,N$, evaluating the loss function for each couple $(j,s_{i})$ and selecting the one with the lowest value of the loss. Problem \eqref{eq:cart} can be solved in $O(pN)$ time, assuming that for each $j$ the splits $s_{i}$ have been presorted before fitting the tree, which can be done in $O(pNlogN)$, and also that the cost of evaluating the loss function is $O(N)$, which holds for the most common losses used in both classification and regression. The overall cost of fitting a balanced tree is $O(pNlogN)$, while in the general case of unbalanced trees it becomes $O(pN^2)$ \citep{book_esl}.\\

When the input data belongs to a functional Hilbert space, like the $L^{2}$ space of square integrable functions, the multivariate approach of splitting the node by iterating through all the $p$ dimensions cannot be used directly, as we first need a preprocessing step in order to estimate the true functions and evaluate them on the same $p$-dimensional grid. In particular, we express $x \in L^{2}(I)$ as a finite expansion on a suitable basis $x(t)= \sum_{j=1}^{J} \xi_{j} \psi_{j}(t)$, where the coefficients $\xi_{j}$, the basis functions $\psi_{j}$ and the number of basis $J$ are estimated from the raw data. If the number of missing values is not excessive, each function can be recovered individually by fixing a basis or using free knots regression splines. Otherwise, mixed-effects models \citep{fda_pcasparse} or local smoothing \citep{fda_sparselongdata} can be used with functional principal components in order to augment each individual sample with dataset-wide information, compensating for the unobserved parts of the domain of the individual sample. Regardless of the estimation approach that is chosen, this is already a viable option that properly manages the problems of noise and missing values, but it's not directly tied to the CART algorithm. While this decoupling can be beneficial is some cases, a tighter integration with the classification/regression methods could lead to better generalization. The tree structure is a natural candidate for this effort, as it is composed of multiple simple elements (the nodes), but at the same time allows for extreme flexibility in dealing with different types of inputs and responses (different splitting rules and loss functions). While a straightforward approach could be repeating smoothing and/or alignment in each node, this would not alleviate the issues related to high dimensional splits and interpretability. Following the idea of looking at the single functional sample as a whole unit, we recover the input functions once before fitting the tree, and instead propose a splitting rule that acts as a variable selection mechanism over subsets of the domain, avoiding to rely exclusively on the single evaluations and therefore reducing the number of splitting dimensions. In particular, by means of functional linear/logistic regression, we estimate multiple data-driven measures that will be used to compute a set of weighted integral features for each input function, with the tree splitting process that is based on such features instead of the single evaluations. As the infinite dimensional functional linear model is ill posed, we adopt the approach of fixing a rich enough basis with a suitable penalty as an identifiability constraint \citep{fda_splineflm,fda_splineerrors,fda_smoothsplines,fda_rkhs}. In particular, we employ the following simple grid basis for the weight function $w$ inside each node: 

\begin{equation*}
  w(t) = \sum_{j=1}^{p} w_{j} \phi_{j}(t)  \hspace{1.5cm}
  \phi_{j}(t) = 
  \begin{cases}
  1 & \hspace{0.1cm} \text{if} \hspace{0.2cm} \frac{j-1}{p} < t \leq \frac{j}{p} \\
  0 & \hspace{0.1cm} \text{otherwise} 
  \end{cases}
\end{equation*}

where $p$ is the length of the estimation grid of the smoothed inputs and is fixed before fitting the tree. Consider the training set $\mathcal{D}=\lbrace (x_i,y_i) \rbrace_{i=1}^{N}$ with random functions $x_i \in L^{2}(I)$ and responses $y_i \in \{1,..,K\}$ for classification or $y_i \in \mathbb{R}$ for regression. Let $w: I \rightarrow \mathbb{R}$ be a  weight function and let $s\in \mathbb{R}$ be a splitting threshold. Let $f: L^{2}(I) \rightarrow \mathbb{R}$ be a feature extractor chosen from a predefined set $\{f_1,..,f_M\}$. Split the input space into two subregions

\begin{center}
$\mathcal{R}_{1}^{f}(w,s) = \lbrace  x | f (x,w) \leq s \rbrace    \:\:\:\:\:\:\:\:\:   \mathcal{R}_{2}^{f}(w,s) = \lbrace  x | f(x,w) > s \rbrace$
\end{center}

that are chosen by solving the following optimization problem:\\

\begin{equation}
\label{eq:nodesplit}
\underset{f,w,s}{\Huge{min}} \: \left[ \mathcal{L}(\mathcal{R}_{1}^{f}(w,s)) + \mathcal{L}(\mathcal{R}_{2}^{f}(w,s)) \right]
\end{equation}\\

where $\mathcal{L}$ is the loss function of the node which depends on the task (classification or regression). There is no restriction on the number of splitting rules that can be used, as one can always iterate through $M$ feature extractors and choose the corresponding rule that results in the best split with respect to the purity criteria of the node. Moreover, the splitting rules can also depend on the task and on the nature of the input data, like in the presence of any combination of categorical, numerical or functional inputs. Motivated by FDA, we propose the following splitting rules for both classification and regression with one functional input:

\begin{center}
$\mathcal{R}_{1}^{\mu}(w,s) = \left\{  x \big{|} \frac{1}{|I|}\int_{I} x(t)w(t)dt \leq s \right\}$  \\ 
$\mathcal{R}_{2}^{\mu}(w,s) = \left\{  x \big{|} \frac{1}{|I|}\int_{I} x(t)w(t)dt > s \right\}$ 
\end{center}

\begin{center}
$\mathcal{R}_{1}^{\sigma^{2}}(w,s) = \left\{  x | \int_{I} (x(t)-\bar{x})^2w(t)dt \leq s \right\}$  \\ 
$\mathcal{R}_{2}^{\sigma^{2}}(w,s) = \left\{  x | \int_{I} (x(t)-\bar{x})^2w(t)dt > s \right\}$ 
\end{center}

\begin{center}
$\mathcal{R}_{1}^{cos\theta}(w,s) = \left\{  x |  \rho_{w}(x,\bar{x}_{node}) \leq s \right\}$  \\ 
$\mathcal{R}_{2}^{cos\theta}(w,s) = \left\{  x | \rho_{w}(x,\bar{x}_{node}) > s \right\}$ 
\end{center}
\vspace{0.5cm}
where $\bar{x}\in\mathbb{R}$ is the mean of the function $x\in L^{2}(I)$ with respect to the measure induced by $w$, so that the first rule splits the node depending on the weighted mean of the function over its domain, while the second rule splits the node with respect to the function's weighted variance over the domain. The third rule is based on the the weighted functional cosine similarity

\begin{center}
$\rho_{w}(x,z) = \frac{\int_{I} x(t)z(t)w(t)dt}{\sqrt{\int_{I} x^{2}(t)w(t)dt}\sqrt{\int_{I} z^{2}(t)w(t)dt}}$ 
\end{center}
\vspace{0.5cm}
with $\bar{x}_{node}\in L^{2}(I)$ being the mean function over the functions in the current node of the tree. In the case of classification with $K$ classes, we add $K$ additional cosine splitting rules, by taking $\bar{x}_{k}\in L^{2}(I)$ as the mean function of the class $k$ inside the current node, instead of $\bar{x}_{node}$. Regardless of the splitting rule(s) that we choose to use, the objective function of Problem \eqref{eq:nodesplit} is nonconvex and piecewise constant. For a given $w$, we can solve this problem by enumeration as in the original CART, by computing the splitting thresholds $s_{i}=f(x_i,w)$ for $i=1,..,N$, $f \in \{f_1,..,f_M\}$ and choosing the split with the highest purity. However, minimizing with respect to $w$ yields a global optimization problem, as it involves finding the optimal oblique split. Given the greedy nature of the tree building procedure, we might not need to find a solution that is close to the true unknown global optimum of the node. Nevertheless, solving this problem with a derivative free approach is computationally prohibitive, especially for deeper trees. To avoid solving Problem \eqref{eq:nodesplit} directly, we propose to learn the weight function first and then compute the optimal weighted split for the current node. This means that each node of the tree will be associated with a data-driven weighted $L^{2}$ space, where the weight function that induces the measure of the space is learned in a supervised fashion, and such (possibly signed) measure is used to compute the integral features that will determine the optimal splitting threshold. As for the $M$ feature extractors $\{f_1,..,f_M\}$, there is no restriction on the number of weight functions $\{w_1,..,w_D\}$ that we can compute in each node, as the optimal one will be selected by iterating over all the $MD$ combinations in Problem \eqref{eq:nodesplit}. It is known that sign constraints can be a powerful tool in linear models \citep{signconstr,nnlstsq}, and for this reason we choose to learn three distinct weight functions $\{w_{pos},w_{neg},w_{sgn}\}$, with positive, negative and no sign constraints, by solving three convex optimization problems that differ by their respective set of constraints. For the task of classification, we solve the following problem: \\

\begin{equation}
\label{eq:classw}
\underset{\scaleto{w_{0},w}{4pt}}{\scaleto{min}{7pt}} \:\: -\sum_{i=1}^{N_{node}} \left[ y_{i} \left( w_{0}+\int_{I}x_{i}(t)w(t)dt \right) - \ln\left(1+{\rm e}^{w_{0}+\int_{I}x_{i}(t)w(t)dt}\right) \right] + \lambda \int_{I} w^{2}(t)dt
\end{equation}
\begin{center}
$ s.t. \:\:\:\:\:\:
\int_{I} w(t)dt=|I| 
\:\:\:\:\:
or
\:\:\:\:
\begin{cases}
\int_{I} w(t)dt=|I| \\
w(t)\geq 0 
\end{cases}
or
\:\:\:\:\:
\begin{cases}
\int_{I} w(t)dt= -|I| \\
w(t)\leq 0 
\end{cases}
$
\end{center}
\vspace{0.5cm}

where $w_{0} \in \mathbb{R}$ is the intercept, $N_{node}$ is the number of curves in the current node and $y_{i} \in \left\{0,1\right\}$ is the binary class label. For multiclass classification with $K$ classes, we select the modal class inside the node and relabel the data as in a one-versus-rest (OvR) setting, but only for the purpose of this problem. Note that we do not resort to the standard OvR scheme of solving $K-1$ problems, as we do not use the linear model for making predictions, but only for learning the weight function of the space, with the tree splitting mechanism that provides for class discrimination. While it is true that solving all the $K-1$ problems could lead to better splits and therefore purer nodes at shallower depths, this would further increase the computational burden, without any guarantee of global optimality for the fully grown tree. The hyperparameter $\lambda \geq 0$ that controls the amount of shrinkage is selected by cross-validation, and in our implementation it is set before fitting the tree, while in principle one could select the optimal one inside each node, without any guarantee that this would increase generalization. In order to have a meaningful comparison between the different regions of the domain, the input curves are standardized over the whole dataset, again only for this step. Analogously, for the task of regression we solve:\\

\begin{equation}
\label{eq:regw}
\underset{\scaleto{w_{0},w}{4pt}}{\scaleto{min}{7pt}} \:\:\: \sum_{i=1}^{N_{node}} \left[ y_{i}  -w_{0} - \int_{I}x_{i}(t)w(t)dt \right]^{2} + \lambda \int_{I} w^{2}(t)dt
\end{equation}
\begin{center}
$ s.t. \:\:\:\:\:\:
\int_{I} w(t)dt=|I| 
\:\:\:\:\:
or
\:\:\:\:
\begin{cases}
\int_{I} w(t)dt=|I| \\
w(t)\geq 0 
\end{cases}
or
\:\:\:\:\:
\begin{cases}
\int_{I} w(t)dt= -|I| \\
w(t)\leq 0 
\end{cases}
$
\end{center}
\vspace{0.5cm}
 
with $y_{i} \in \mathbb{R}$. Therefore, the difference between Problems \eqref{eq:classw} and \eqref{eq:regw} is in the first term of the objective function, which is either a functional logistic loss or a functional ordinary least squares. In the context of FDA, the usual choice is often a roughness penalty rather than the ridge penalty that we employ, as the intent is to estimate a smoother linear model. However, we do not employ these models to directly perform the split, but rather use them to learn the measure associated with the weighted $L^{2}$ space of the node, with the sets of constraints that are such that $\mu(I) = \nu (I)$, with $\mu$ the Lebesgue measure and $\nu$ the measure induced by the resulting weight function. In particular, the weight function has to be positive in order to define a weighted $L_{2}$ space and have statistically well posed feature extractors $f_m$. For this reason, we take the absolute value of the weight function $w_{neg}$, which does not change the relative importance of the different parts of the domain $I$. Unfortunately, this interpretation breaks down in the case of the signed measure $w_{sgn}$, where the weighted cosine similarity is not defined, but we still retain the first two splitting rules, which are both linear models. Without solving an additional optimization problem, we also include the uniform weight function in order to extract the unweighted features, as in principle there is no guarantee that the learned representations will be beneficial. On a side note, the type of penalty in the optimization problem could either be treated as an hyperparameter to be selected by cross-validation, or multiple penalties could be included inside each node, further increasing the number of learned weight functions $D$, together with the training time. Overall, our approach could be seen as an oblique tree where the linear combination determines the regions of the domain $I$ that are most informative with respect to the task, and the splitting values are obtained as a weighted function of the input curves. Another analogy could be drawn with neural networks, as our approach is also similar to having a single neuron for each node of the tree, with one of the feature extractors as the adaptively chosen activation function.

From the computational standpoint, suppose that the true input functions have been estimated and the evaluation grid is of length $p$. For each node of the tree, we first need to solve either Problem \eqref{eq:classw} for classification or Problem \eqref{eq:regw} for regression, once for each of the $D$ sets of constraints. Both problems are convex with linear equality and inequality constraints and can be solved with interior point methods in $O(\sqrt{p})$ iterations \citep{iposurvey}, using Newton's method at a cost of $O(p^3)$ for a single iteration. To compute the split, for each of the $D$ resulting weight functions ($D+1$ if we include the uniform one), we first need to compute the $M$ weighted features for the $N_{node}$ curves in the current node at a cost of $O(pDMN)$, and then we solve Problem \eqref{eq:nodesplit}, by sorting the splits in $O(DMNlogN)$ time and finding the optimal one in $O(DMN)$, again assuming that the loss function can be evaluated in $O(N)$. As the splits depend on the current $w_{d}$ in each node, we cannot presort them once before fitting the tree. Given that the number of features $M$ and the number of weight functions $D$ are independent from $p$ and $N$ (and usually much smaller), the asymptotic cost of splitting a single node is $O(NlogN + p^{3.5})$. In the case of balanced trees, the asymptotic cost of fitting a tree is $O(Nlog^{2}N + p^{3.5}logN)$, which becomes $O(N^{2}logN + Np^{3.5})$ in the unbalanced case.

\section{Experimental Results}
\label{experiments}
In the following sections, we show a classification simulation study with two functional covariates and multiple real world applications for both multiclass classification and regression. In order to validate the performance of our approach ($\mu$CART), we compare it against known tree-based methods like CART, RF and GBT. For datasets that are not pre-smoothed, we recover the true functions by means of free knots regression splines and evaluate them on the same $p$-dimensional equispaced grid. This preprocessing step is a standard practice in FDA and is shared between all methods, allowing for a meaningful comparison. Moreover, we also test CART, RF and GBT with an additional preprocessing approach (FE), which consists in extracting the same features that we implement in our nodes, with the objective of showing the effectiveness of learning node-wise representations. The code has been implemented in python3 and all the known methods and cross-validation routines rely on Scikit-learn \citep{scikit}. The optimization problems have been modeled with Pyomo \citep{book_pyomo} and solved with Ipopt \citep{ipopt}. All the computation has been done on a desktop CPU with 4 cores at 4.4GHz and 16GB of RAM.

\subsection{Classification}
For the task of classification, we first show a synthetic problem with two functional covariates and class membership that depends on their shape in a subset of the domain. The second case study deals with infrared spectra classification for product authentication in chemometrics, while the third application is from the medical field and is about (multiclass) disease classification from gut microbiota. The growth of all trees is controlled by limiting the minimum number of samples required to be at a leaf node, with Gini index as the function to measure the quality of the split for CART, $\mu$CART and RF, while for GBT the loss function is the deviance and the quality of the split is measured by the improved least squares criterion \citep{boosting}. The number of trees for the ensemble methods has been fixed to 150, while for RF in particular, the bootstrap sampling is class balanced and the maximum number of features is chosen adaptively between $\left\{ all,log,sqrt \right\}$. For all case studies, we report the accuracy on the test set with the corresponding standard deviation, averaged over five random repetitions (only one for the simulation) of 5-fold cross-validation with 3-fold cross-validation for grid search hyperparameter optimization. Moreover, we also report the average tree height over the folds but only for $\mu$CART and CART, as the other methods are not easily interpretable and the mean tree height of an ensemble of trees is arguably not as informative.

\subsubsection{Simulation Study}
\begin{figure}[H]
  \centering
    \begin{subfigure}[c]{0.49\textwidth}
      \includegraphics[width=\textwidth]{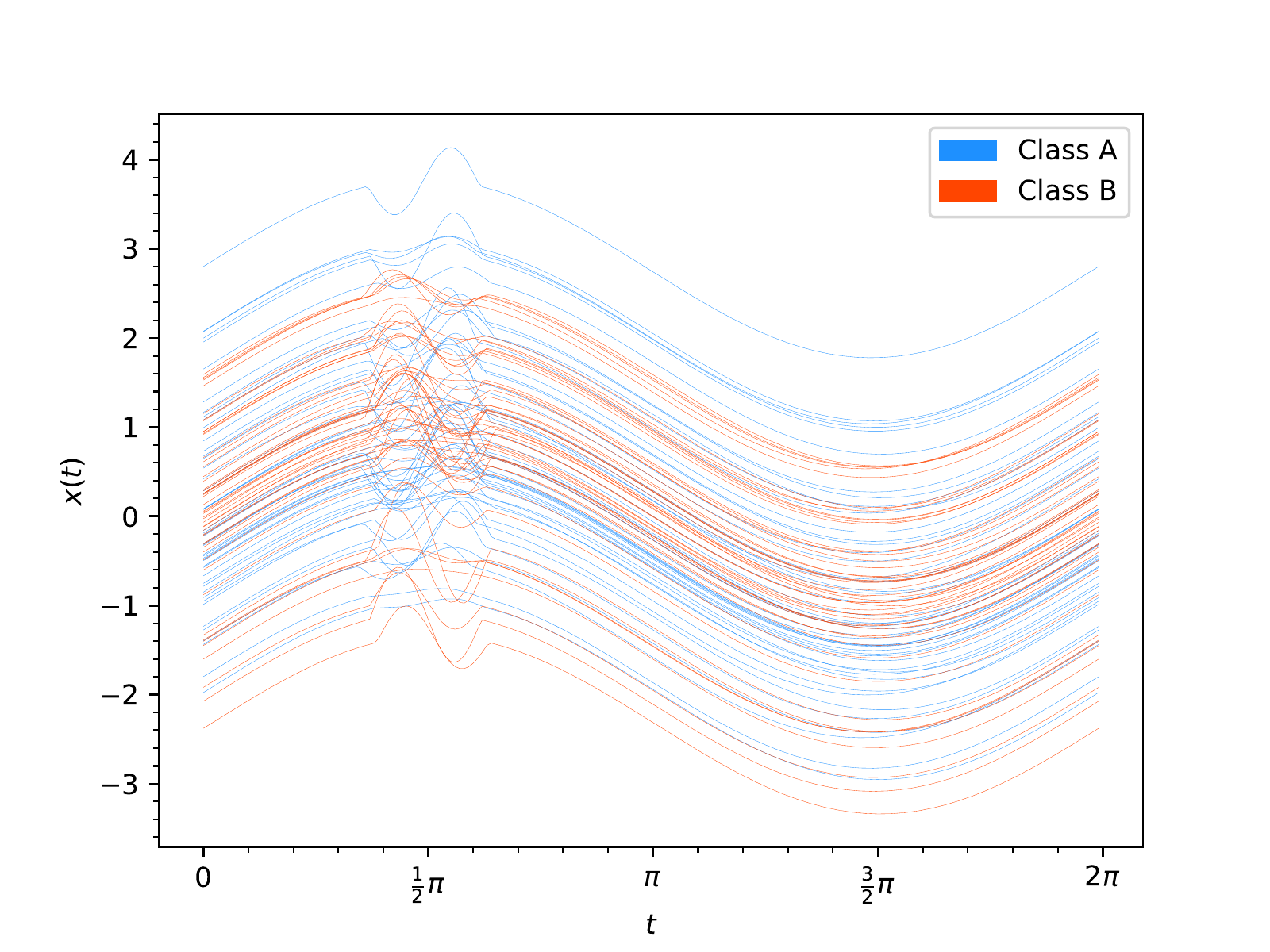}
      \caption{First Functional Covariate}
    \end{subfigure}
    \begin{subfigure}[c]{0.49\textwidth}
      \includegraphics[width=\textwidth]{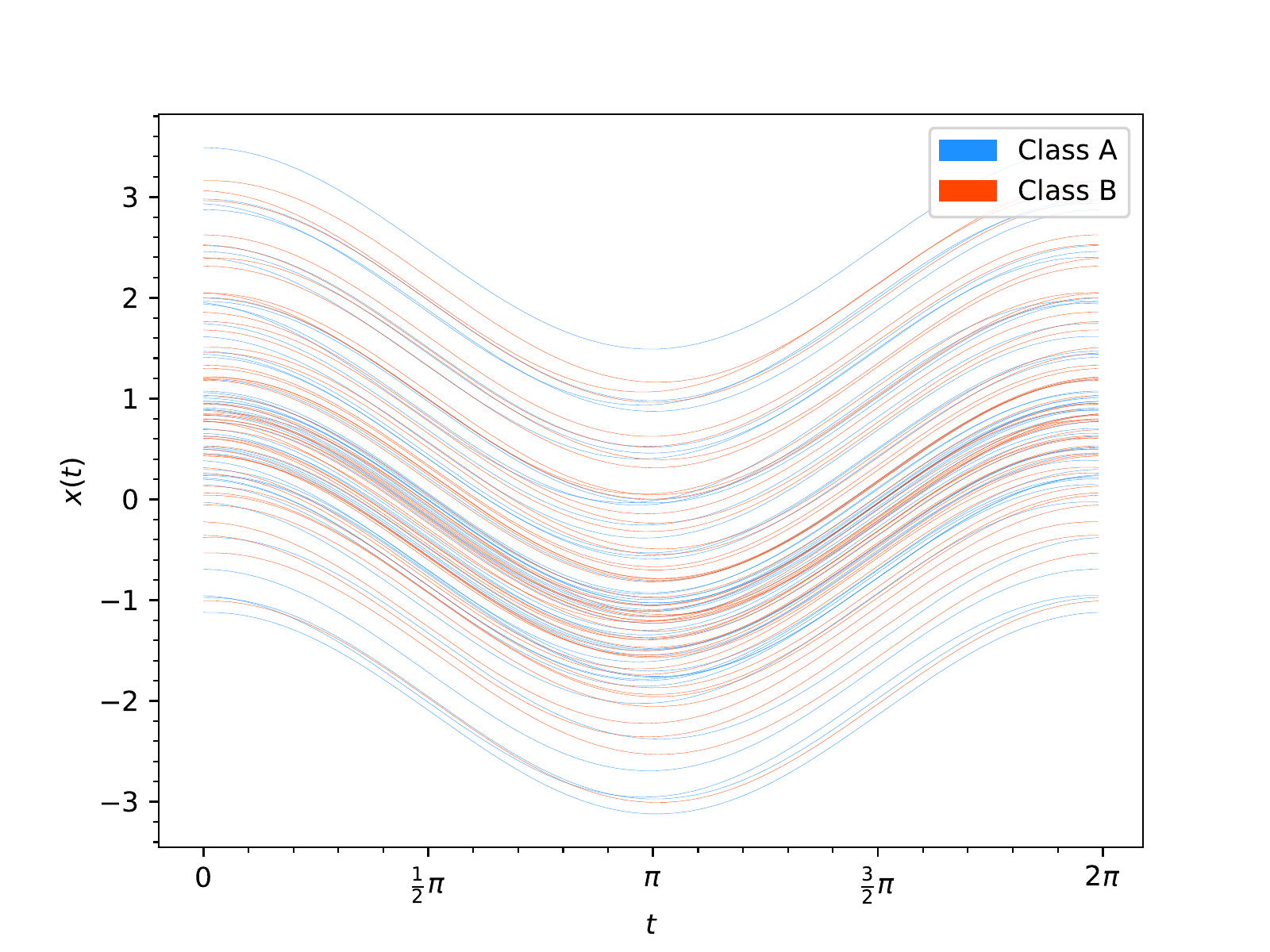}
      \caption{Second Functional Covariate}
    \end{subfigure}
  \caption{Simulated data, classification}
  \label{fig:csim}
\end{figure}

This simulation is a binary classification problem with two functional covariates, in which the information to discriminate between the classes is localized in the first part of the domain of the first covariate, while the second one is not informative, as shown in Figure \ref{fig:csim}. Let $i=1,..,N$ with $N$ the sample size (each sample is a curve) and let $j=1,..,p$ with $p$ the number of points of the discrete evaluation grid over the domain. Regarding the first covariate, the base functional model is a sine curve evaluated in $t_{j} \in [ 0,2\pi ]$, with a curve dependent amplitude term $b_{1i} \sim \mathcal{N}(0,1)$ and a curve dependent phase term $\phi_{1i} \sim \mathcal{N}(0,\frac{\pi}{75})$. For $t_{j} \in ( \frac{3}{8}\pi -\phi_{1i},\frac{5}{8}\pi -\phi_{1i})$, we add to the base model another sine that is multiplied by a curve dependent term $\alpha_{i} \sim \mathcal{N}(0,0.3)$, with the sign of $\alpha_{i}$ determining the class. The second covariate is a cosine evaluated in $t_{j} \in [ 0,2\pi ]$, with a curve dependent amplitude term $b_{2i} \sim \mathcal{N}(0,1)$ and a curve dependent phase term $\phi_{2i} \sim \mathcal{N}(0,\frac{\pi}{75})$, but no information on the class membership. The simulation has been repeated with $N$=$\{100,150,200\}$ and the functions have been evaluated in $p$=200 linearly spaced points in $[ 0,2\pi ]$.\\

\begin{flushleft}
$
\begin{cases}
y_i = sgn (\alpha_{i}) \\
x_{1i}(t_{j}) = b_{1i} + \sin(t_{j} + \phi_{1i}), & t_{j} \in [ 0,\frac{3}{8}\pi -\phi_{1i} ] \cup [ \frac{5}{8}\pi -\phi_{1i},\pi ]\\
x_{1i}(t_{j}) = b_{1i} + \sin(t_{j} + \phi_{1i}) + \alpha_{i}\sin(8(t_{j} + \phi_{1i})), & t_{j} \in ( \frac{3}{8}\pi -\phi_{1i} ,\frac{5}{8}\pi -\phi_{1i}) \\
x_{2i}(t_{j}) = b_{2i} + \cos(t_{j} + \phi_{2i}), & t_{j} \in [ 0, 2\pi ]\\
\end{cases}
$
\end{flushleft}

\begin{figure}[H]
\centering
\scalebox{1.2}{
\begin{forest}
[\scalebox{0.23}{\includegraphics{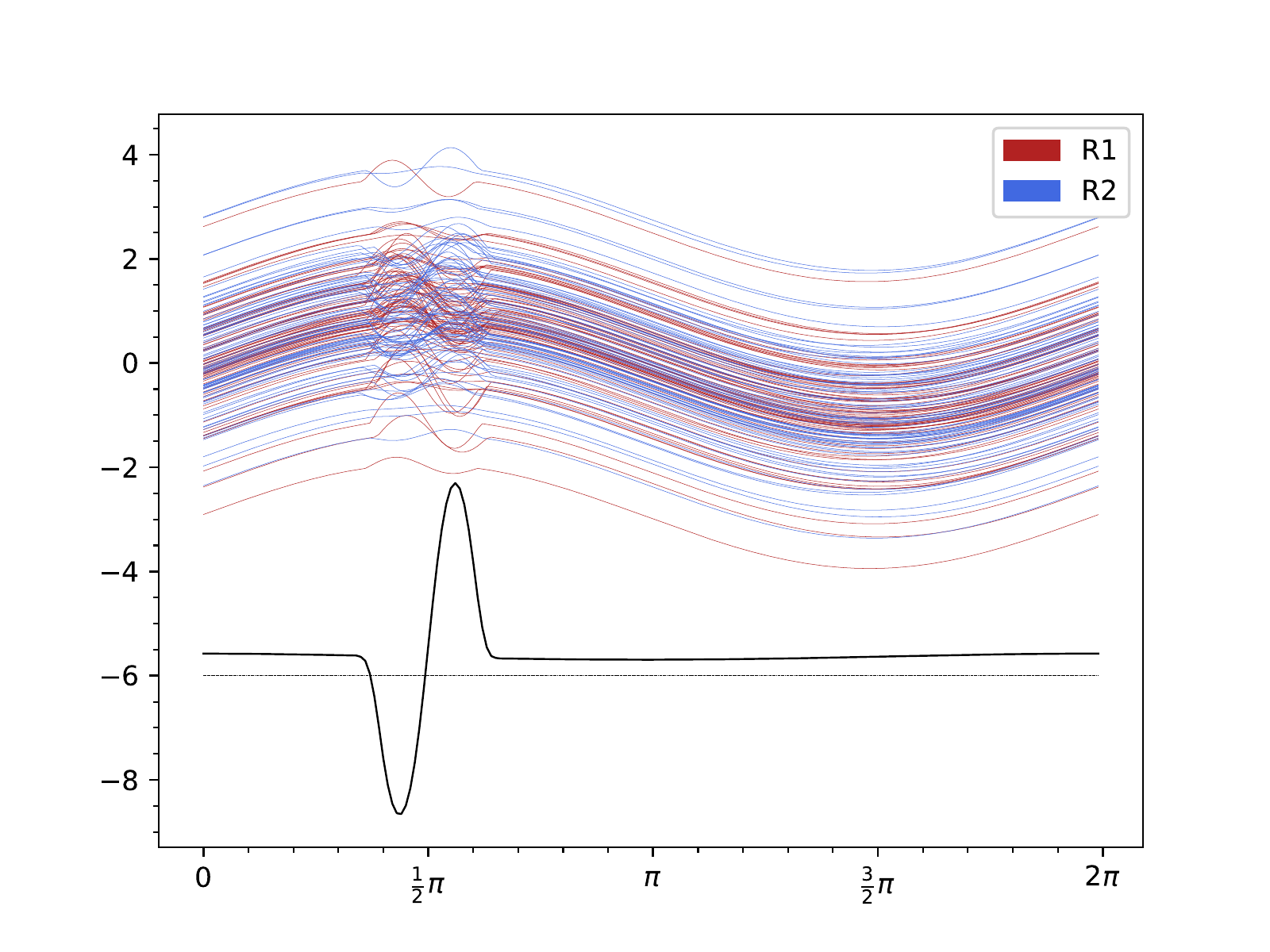}} \\ $w_{sgn} - f_{\sigma^{2}}$={node[midway,below,font=\scriptsize]{1}}
 [\scalebox{0.23}{\includegraphics{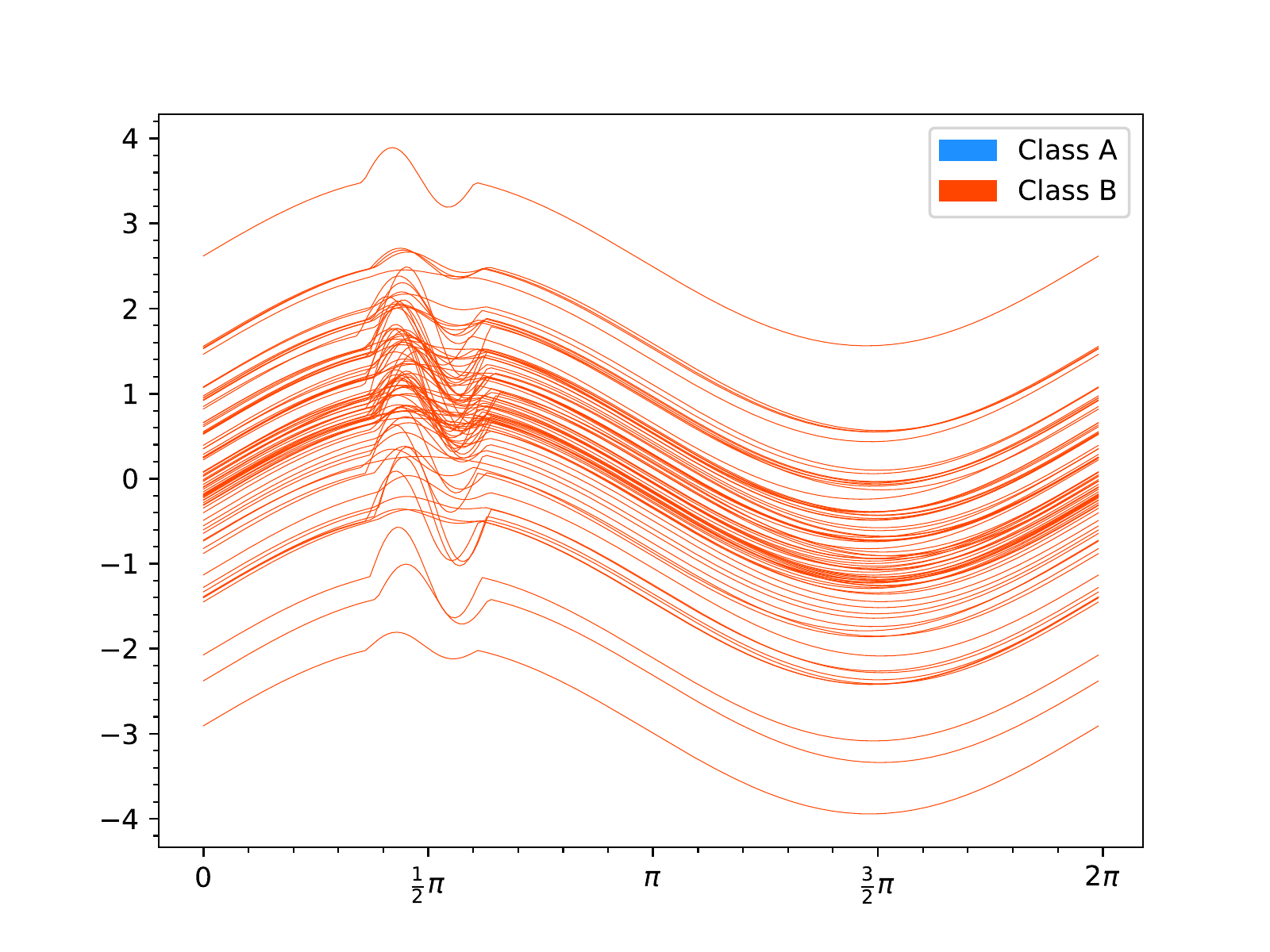}}]
 [\scalebox{0.23}{\includegraphics{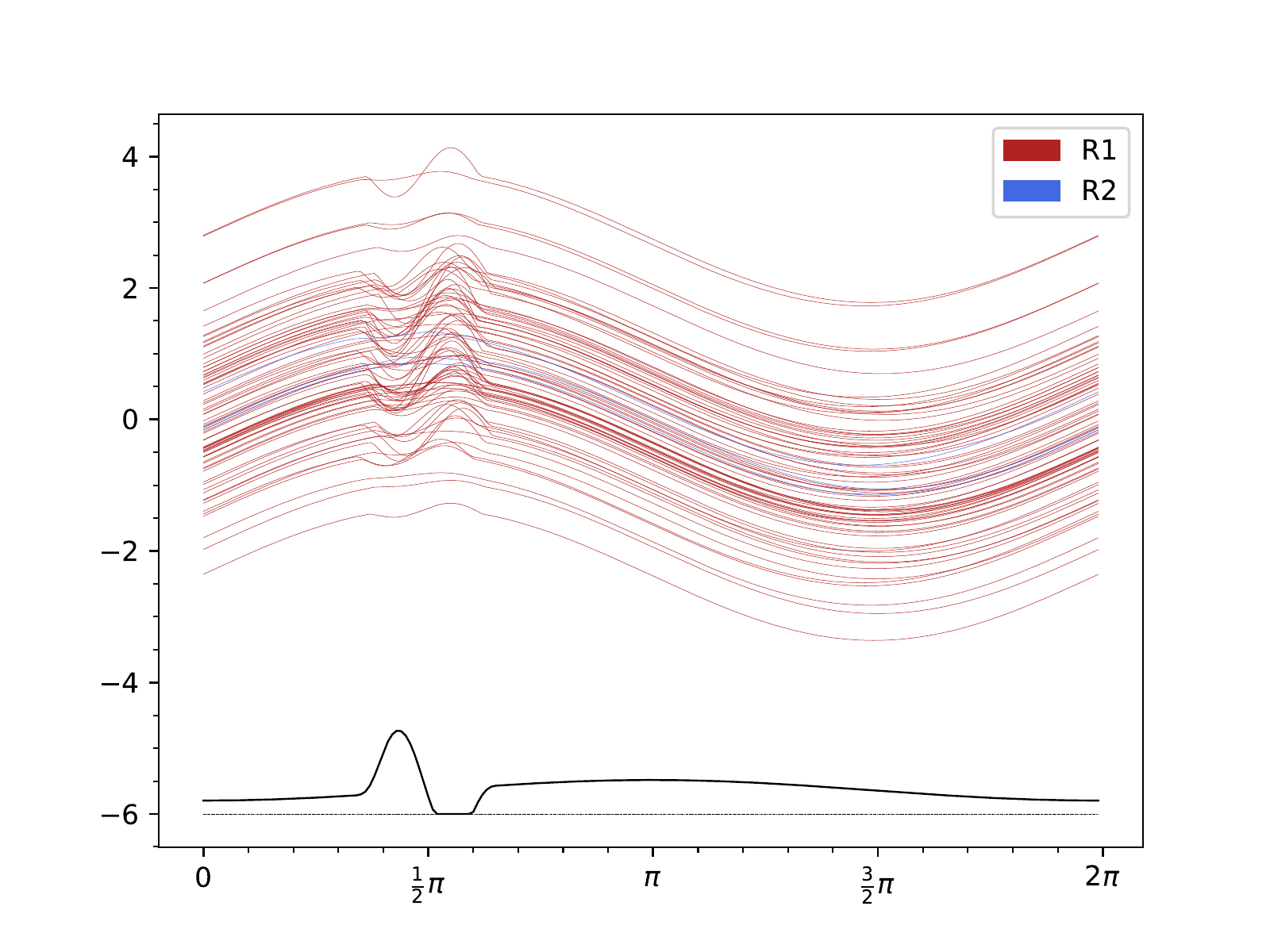}}  \\ $w_{pos} - f_{\sigma^{2}}$={node[midway,below,font=\scriptsize]{1}}
  [\scalebox{0.23}{\includegraphics{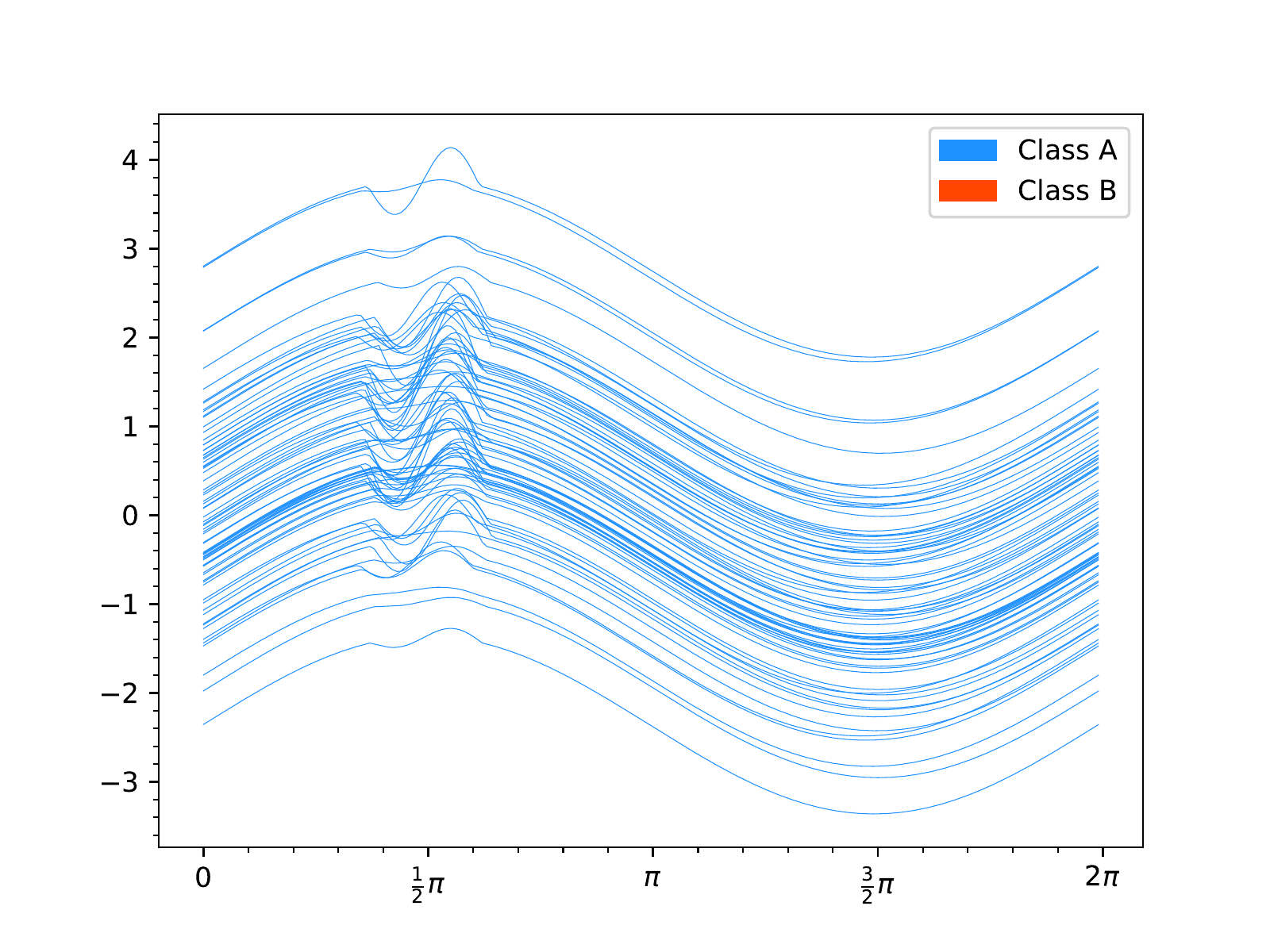}}]
  [\scalebox{0.23}{\includegraphics{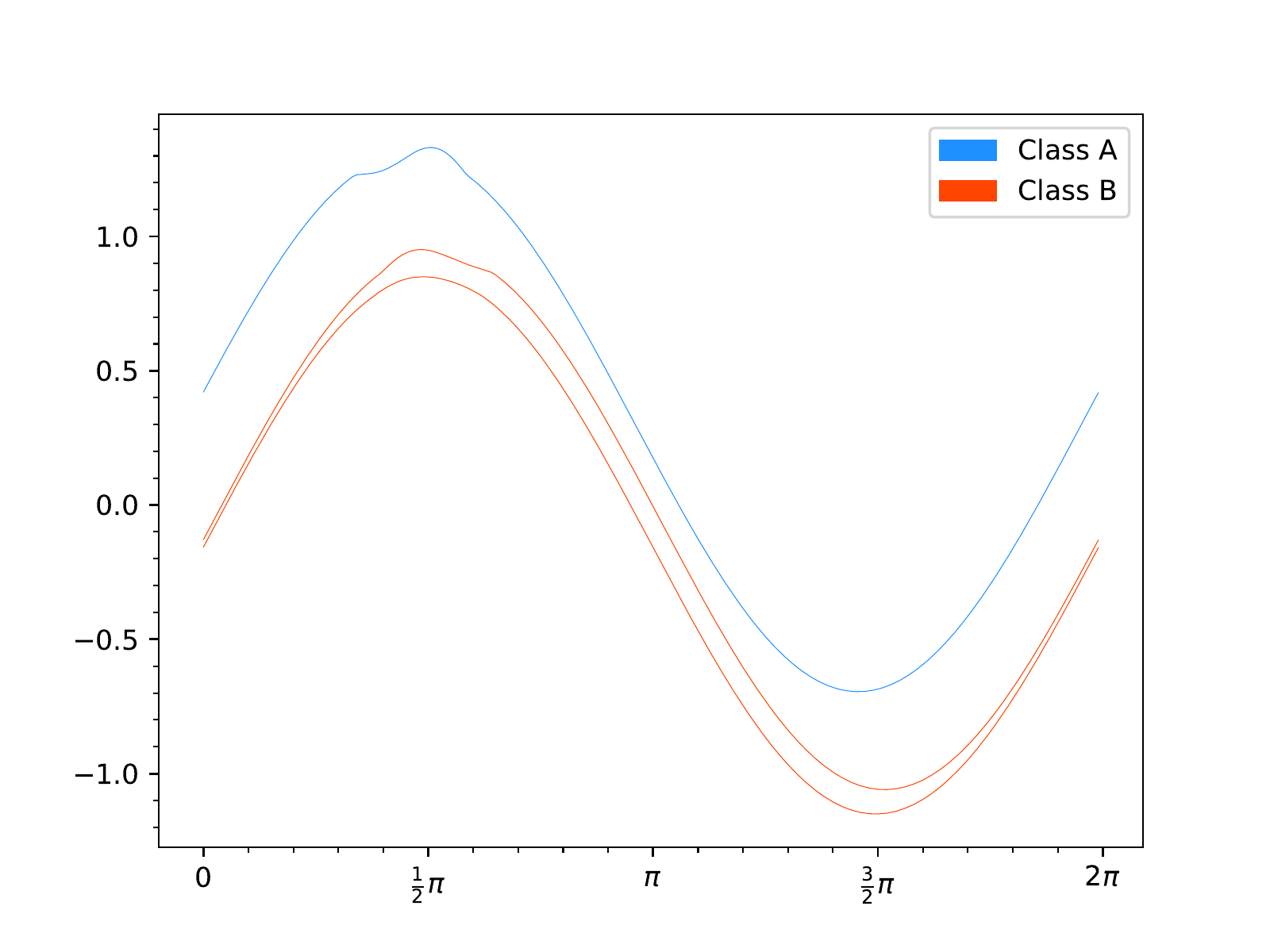}}]
 ]
]
\end{forest}
}
\caption{Simulation study: fully grown tree}
\label{fig:csim_fulltree}
\end{figure}

\vspace{0.5cm}
Table \ref{tab:csim} shows the classification results for all combinations of methods, preprocessing approaches and sample sizes that we tested, together with the average tree heights when applicable. Our proposed tree $\mu$CART has the highest accuracy, showing that the method is able to learn local representations that are effective for the task at hand, as extracting the same features on the whole domain does not allow the other classifiers to have significantly better scores than random guessing, which is intended by design. Using the discretized function evaluations allows the other methods to at least scale with the increasing sample size, but even with double the amount of data and ensembles, the highest accuracy is still lower than the one resulting from $\mu$CART with $N$=$100$. Moreover, our method has the lowest average tree height, which can be beneficial for interpretability, as shown in Figure \ref{fig:csim_fulltree}, where a fully grown tree is depicted, together with the (rescaled) learned measures in the inner nodes. In particular, both inner nodes split with respect to the first functional covariate, the root selects $w_{sgn}$ with the weighted variance as splitting feature ($f_{\sigma^{2}}$), while the other inner node also splits with respect to $f_{\sigma^{2}}$ but selects $w_{pos}$.

\begin{table}[H]
\centering
\caption{\label{tab:csim} Simulation results: classification accuracy and tree height}
\scalebox{0.83}{
\begin{tabular}{l D{,}{\, \pm \,}{-1} D{,}{\, \pm \,}{-1} D{,}{\, \pm \,}{-1} D{,}{\, \pm \,}{-1} D{,}{\, \pm \,}{-1} D{,}{\, \pm \,}{-1}}
\toprule
\midrule
           
 & \multicolumn{3}{c}{Accuracy \%} & \multicolumn{3}{c}{Tree Height} \\
\midrule

 & \multicolumn{1}{c}{$N$=100} & \multicolumn{1}{c}{$N$=150} & \multicolumn{1}{c}{$N$=200} & \multicolumn{1}{c}{$N$=100} & \multicolumn{1}{c}{$N$=150} & \multicolumn{1}{c}{$N$=200} \\
           
\midrule

CART        & 72.2,6   &  80.1,6    &   84.1,6   & 4.4,1  &  5.2,1.2    & 6,0.6   \\

CART+FE     & 53.9,10  &  53.7,13   &   57.5,7   & 5.2,3.3  &  2.2,0.7  & 9.8,3.9    \\

RF          & 76,7     &  82.1,5    &   83.6,5   &   -   &   -  &  -   \\
  
RF+FE       & 54.1,7   &  55.6,11   &   56,9     &   -   &   -  &  -   \\
  
GBT         & 80.8,7   &  84.1,6    &   85.6,3   &   -   &   -  &  -   \\

GBT+FE      & 53.7,4   &  52.1,12   &   60,3     &   -   &   -  &  -   \\

$\mu$CART   & 85.9,18  &  97.4,4    &   98,1     & 1.4,0.5  &  1.6,0.8  & 1.8,0.4    \\
  
\midrule
\bottomrule  
\end{tabular}}
\end{table}

\newpage

\subsubsection{Coffee FTIR Spectra}
\begin{figure}
\begin{center}
\includegraphics[scale=0.8]{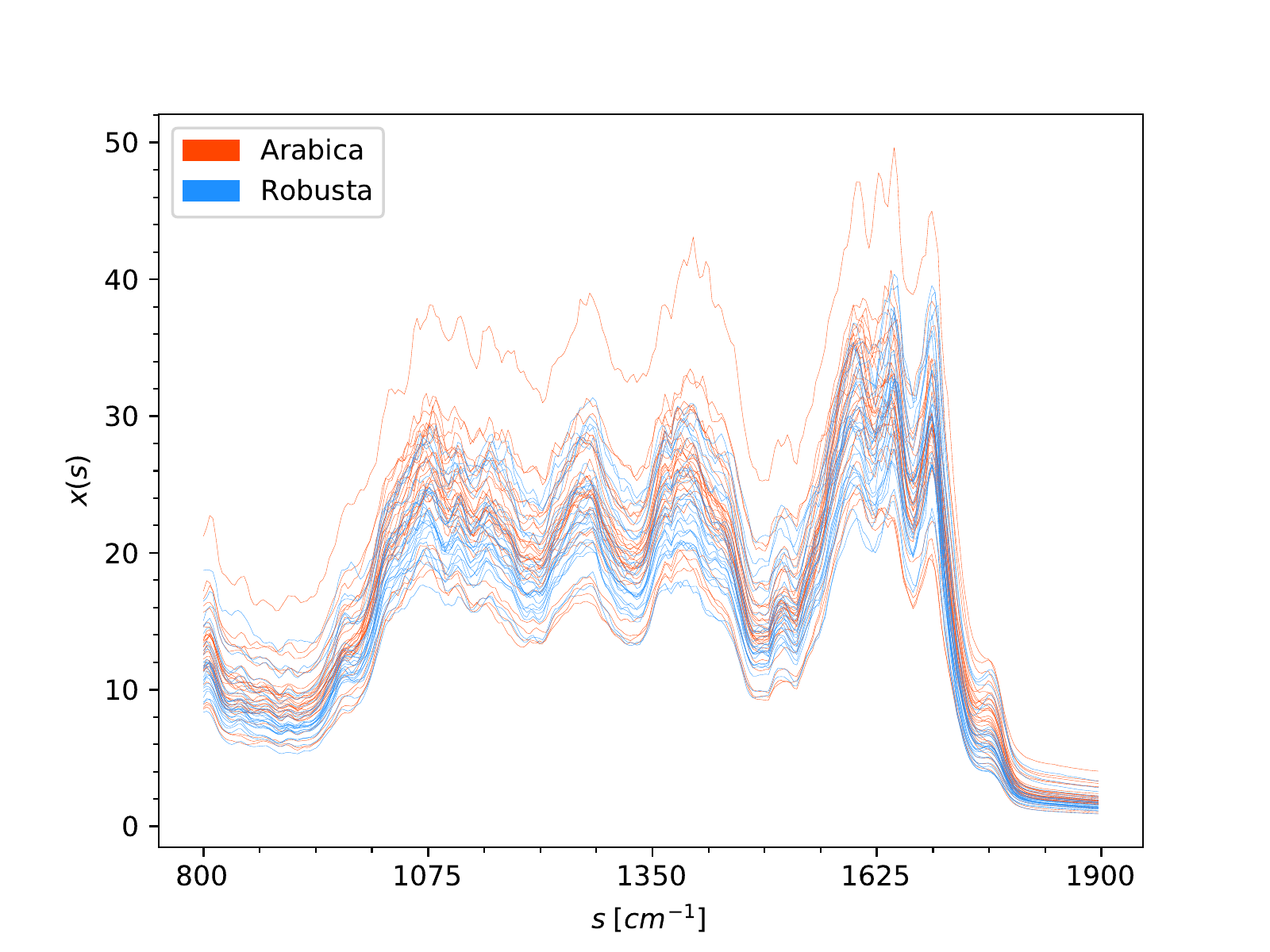}
\caption{\label{fig:coffee} FTIR spectra of coffee samples}
\end{center}
\end{figure}

\begin{figure}[H]
\centering
\scalebox{1.2}{
\begin{forest}
[\scalebox{0.3}{\includegraphics{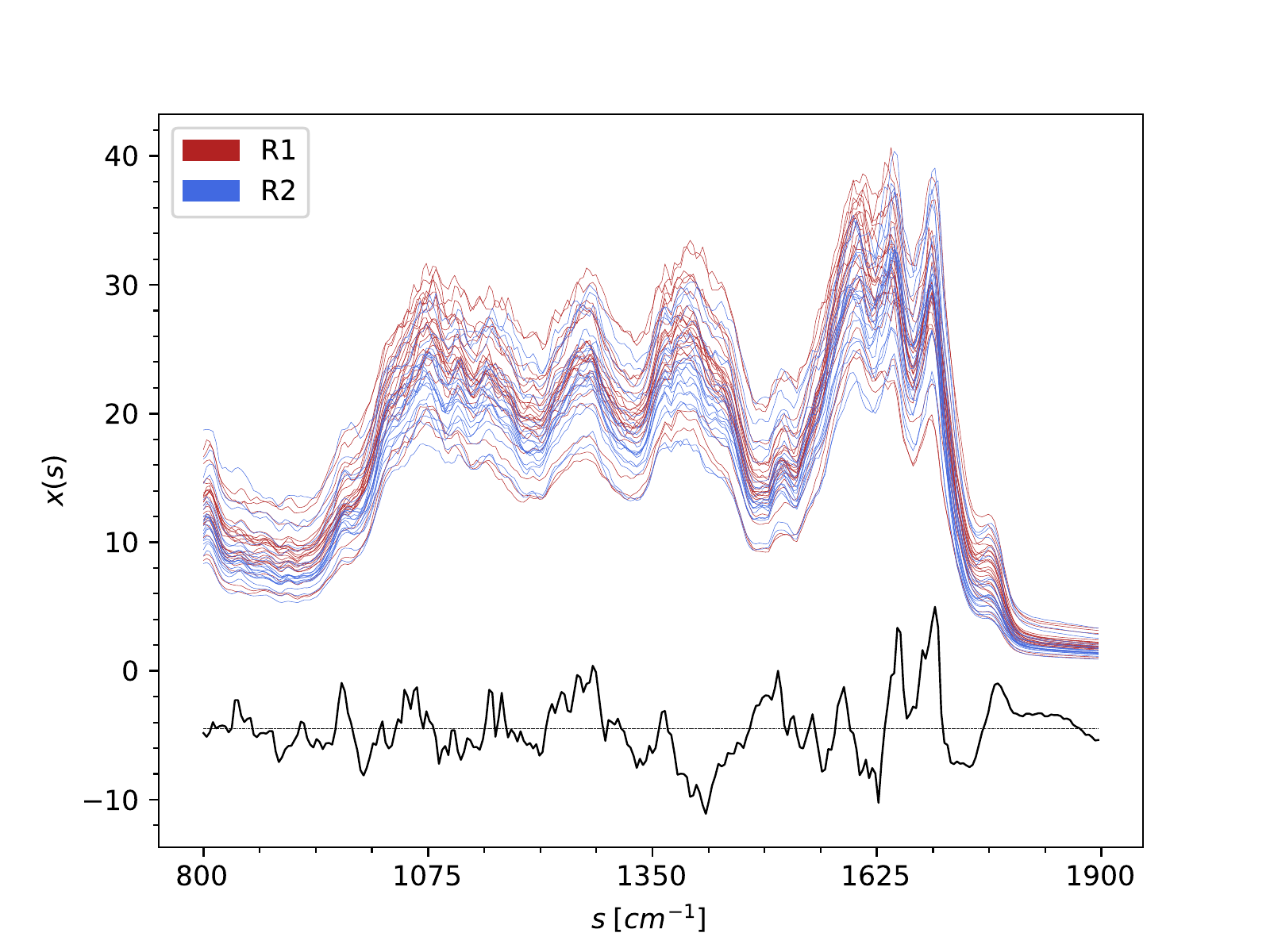}} \\ $w_{sgn} - f_{\mu}$={node[midway,below,font=\scriptsize]{1}}
 [\scalebox{0.3}{\includegraphics{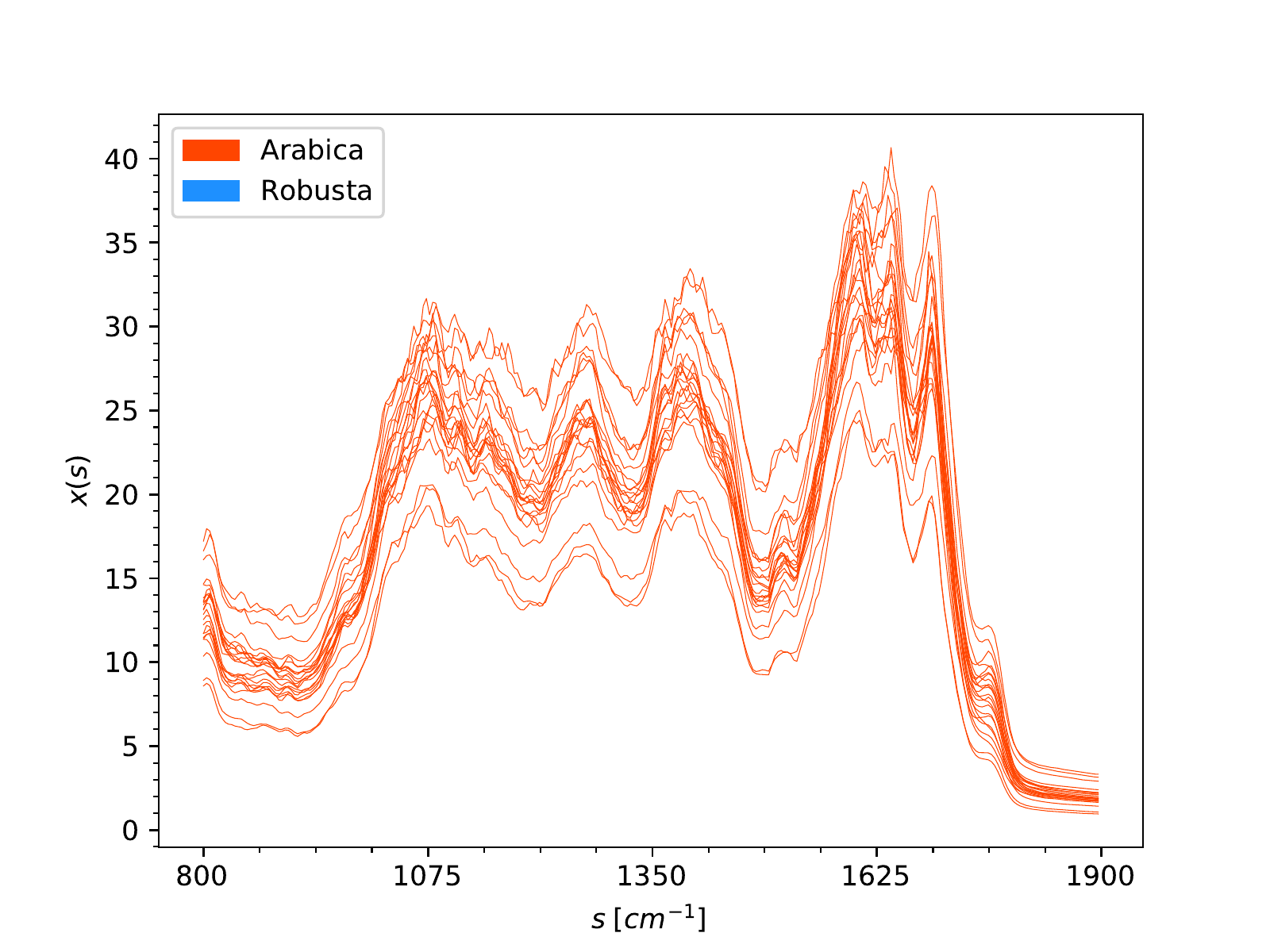}}]
 [\scalebox{0.3}{\includegraphics{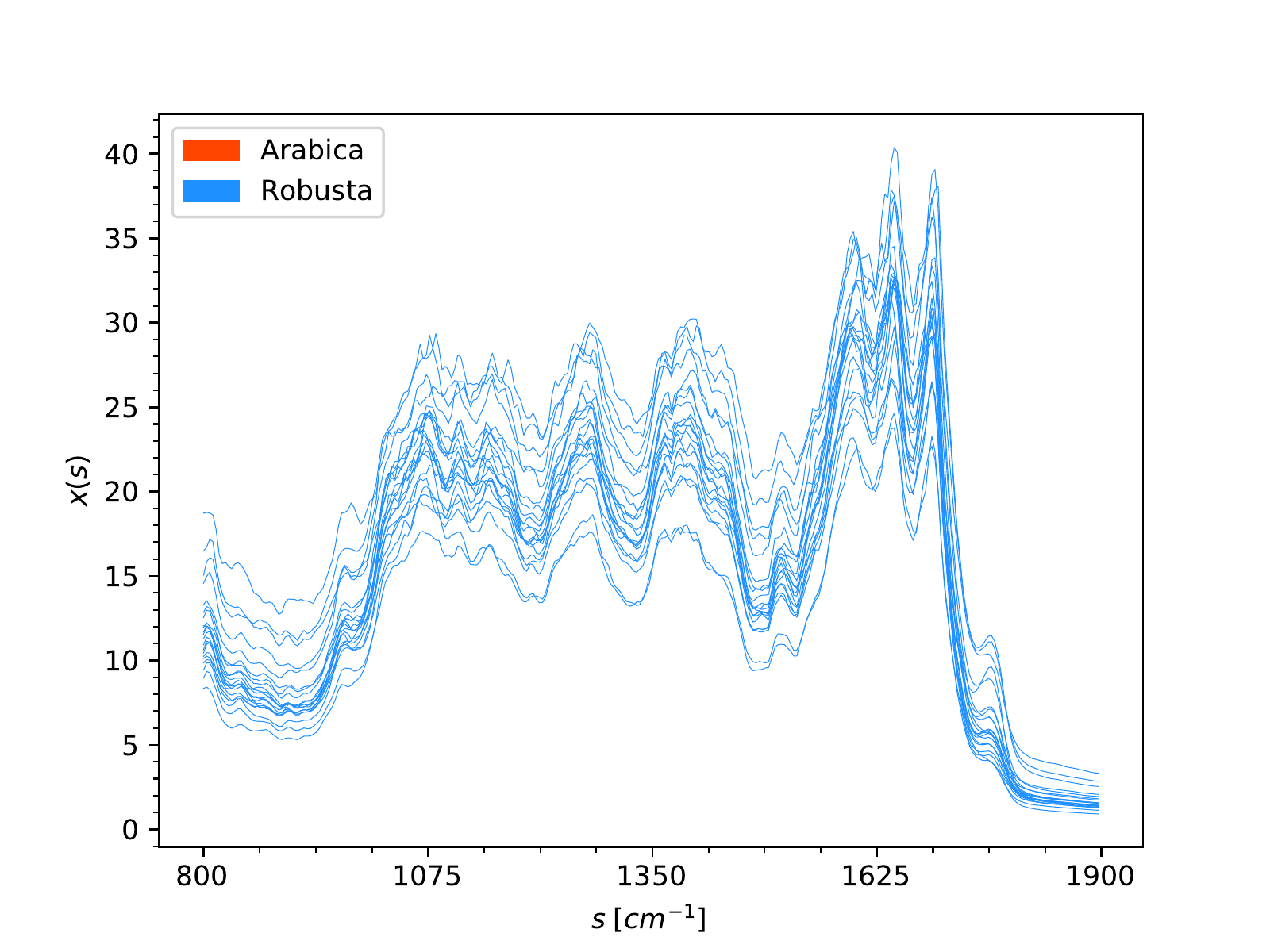}}]
]
\end{forest}
}
\caption{Coffee FTIR Spectra: fully grown tree}
\label{fig:coffee_fulltree}
\end{figure}

This dataset is composed of $N$=56 functional samples of length $p$=286 obtained by Fourier transform infrared (FTIR) spectroscopy with diffuse reflectance (DRIFT) sampling of dried instant coffee. The problem is to discriminate between two types of coffee, \textit{Coffea Arabica} and \textit{Coffea Canephora Robusta}, which have different trade values. The dataset is shown in Figure \ref{fig:coffee} and it is available at \url{http://csr.quadram.ac.uk/wp-content/uploads/FTIRSpectraInstantCoffee.zip}. The raw data has no missing values and negligible sensor noise \citep{coffee}, therefore there is no need to employ further smoothing. Table \ref{tab:coffee} shows the results and the average tree heights, where this time the FE preprocessing is more effective than the discretized function evaluations, with the weighted features of $\mu$CART providing higher accuracy and trees with a single internal node, as shown in Figure \ref{fig:coffee_fulltree}.

\begin{table}[H]
\centering
\caption{\label{tab:coffee} Coffee FTIR Spectra results: classification accuracy and tree height}
\scalebox{1.2}{
\begin{tabular}{l D{,}{\, \pm \,}{-1} D{,}{\, \pm \,}{-1}}
\toprule
\midrule
           
 & \multicolumn{1}{c}{Accuracy \%} & \multicolumn{1}{c}{Tree Height} \\
\midrule
         
\midrule

CART        & 70.4,13  &   2.5,1.1   \\

CART+FE     & 82.9,10  &   3,0.7     \\

RF          & 75.5,14  &   -         \\
   
RF+FE       & 87.3,10  &   -         \\
  
GBT         & 76.9,10  &   -         \\

GBT+FE      & 89.7,7   &   -         \\

$\mu$CART   & 94.3,8  &  1.1,0.1     \\
  
\midrule
\bottomrule  
\end{tabular}}
\end{table}

\subsubsection{Gut Microbiota DGGE}
\begin{figure}
\begin{center}
\includegraphics[scale=0.8]{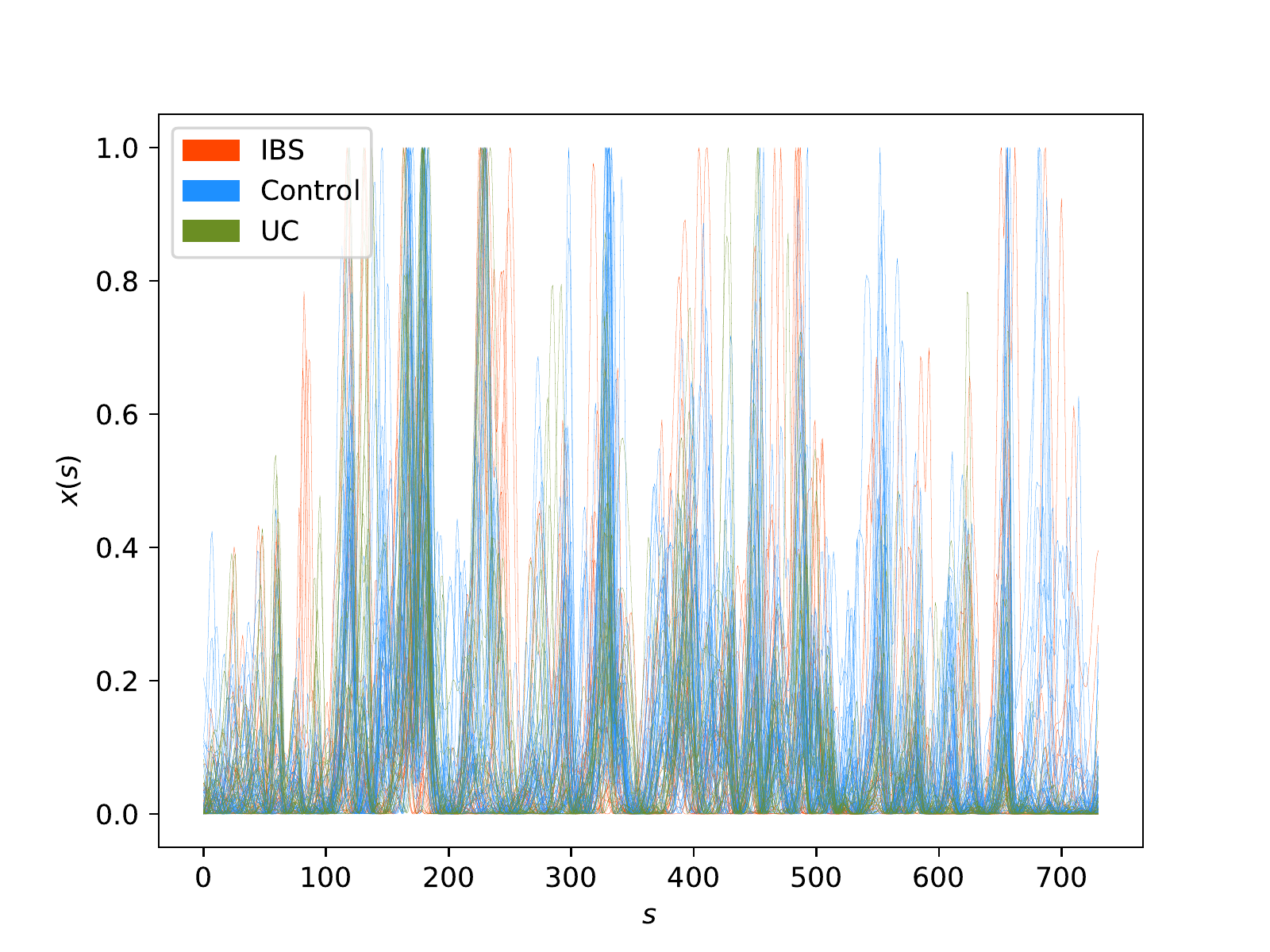}
\caption{\label{fig:dgge} Gut Microbiota DGGE by class: ulcerative colitis (UC), irritable bowel syndrome (IBS) and healthy controls}
\end{center}
\end{figure}

\begin{figure}[H]
\centering
\scalebox{0.52}{
\begin{forest}
[\scalebox{0.47}{\includegraphics{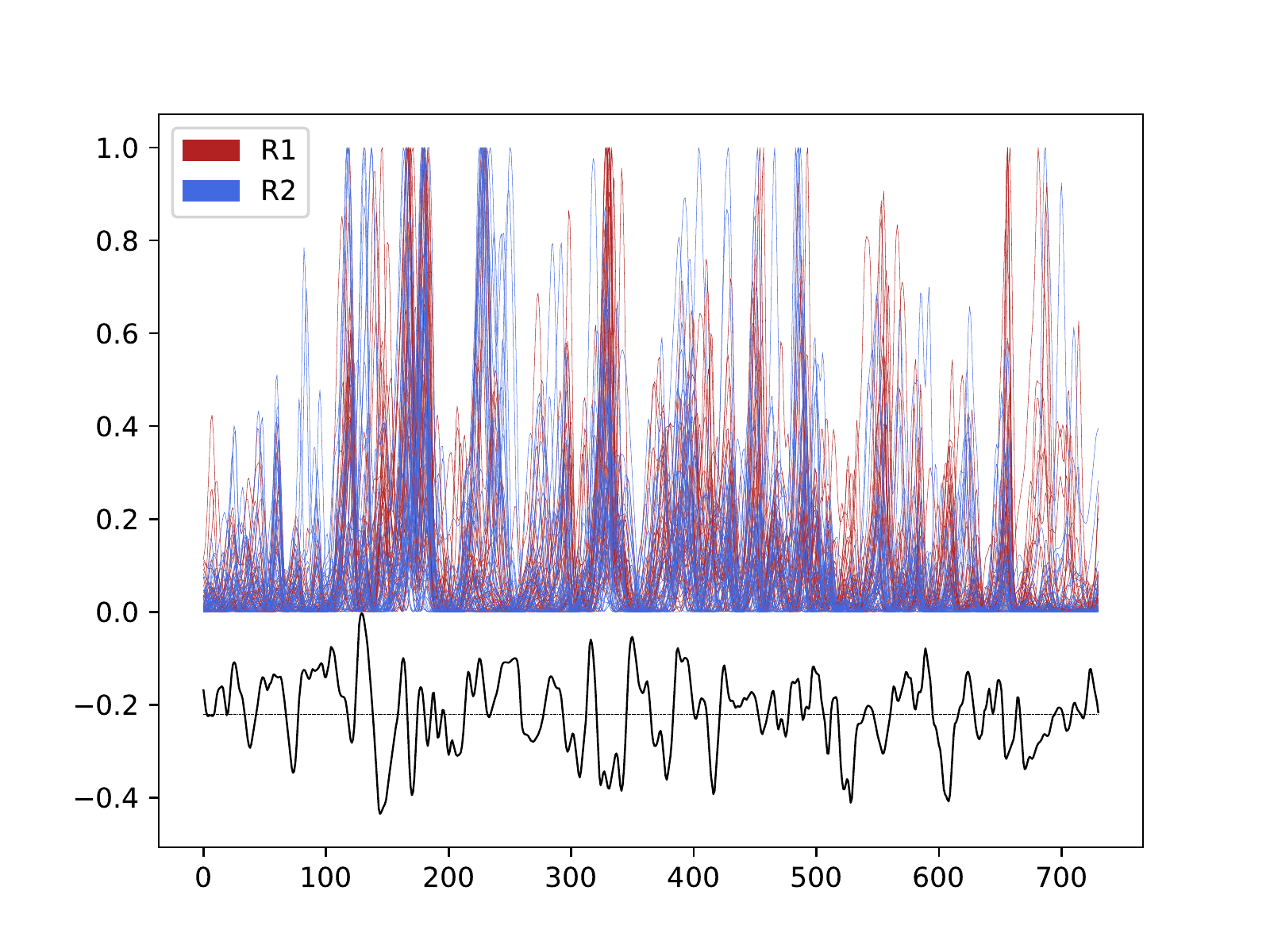}}   \\ $w_{sgn} - f_{\mu}$={node[midway,below,font=\scriptsize]{1}}
 [\scalebox{0.47}{\includegraphics{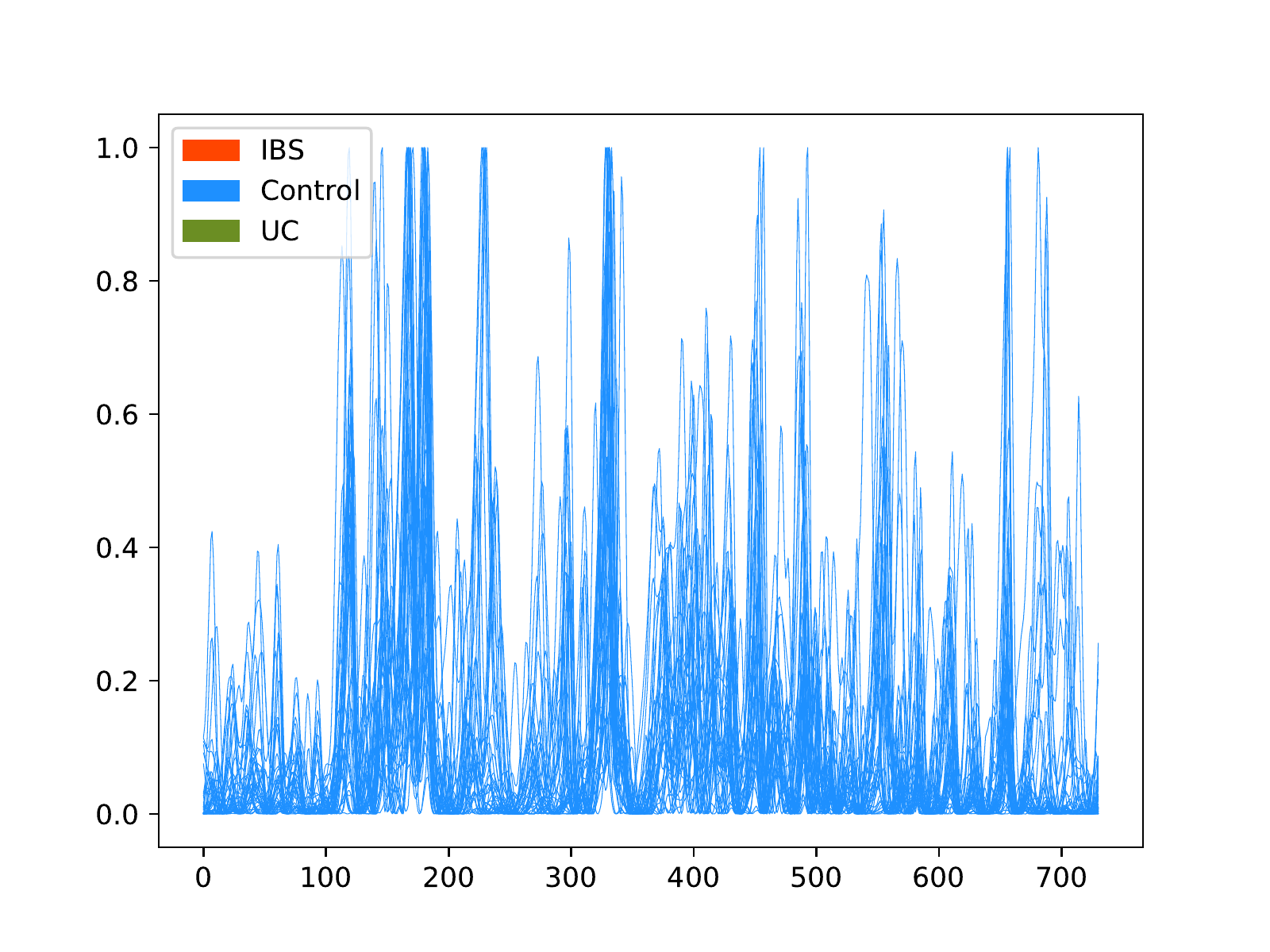}}]
 [\scalebox{0.47}{\includegraphics{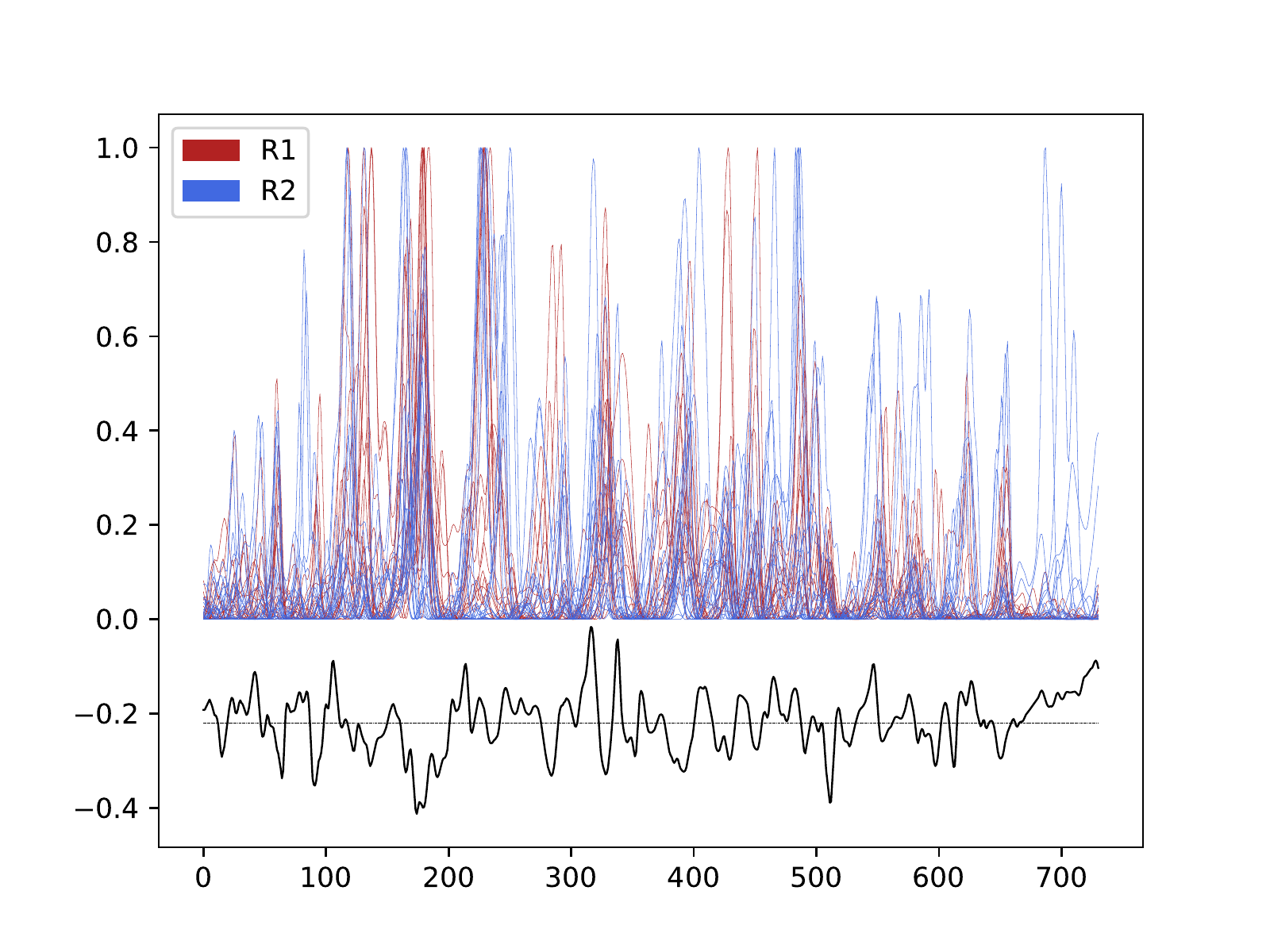}}    \\ $w_{sgn} - f_{\mu}$={node[midway,below,font=\scriptsize]{1}}
  [\scalebox{0.47}{\includegraphics{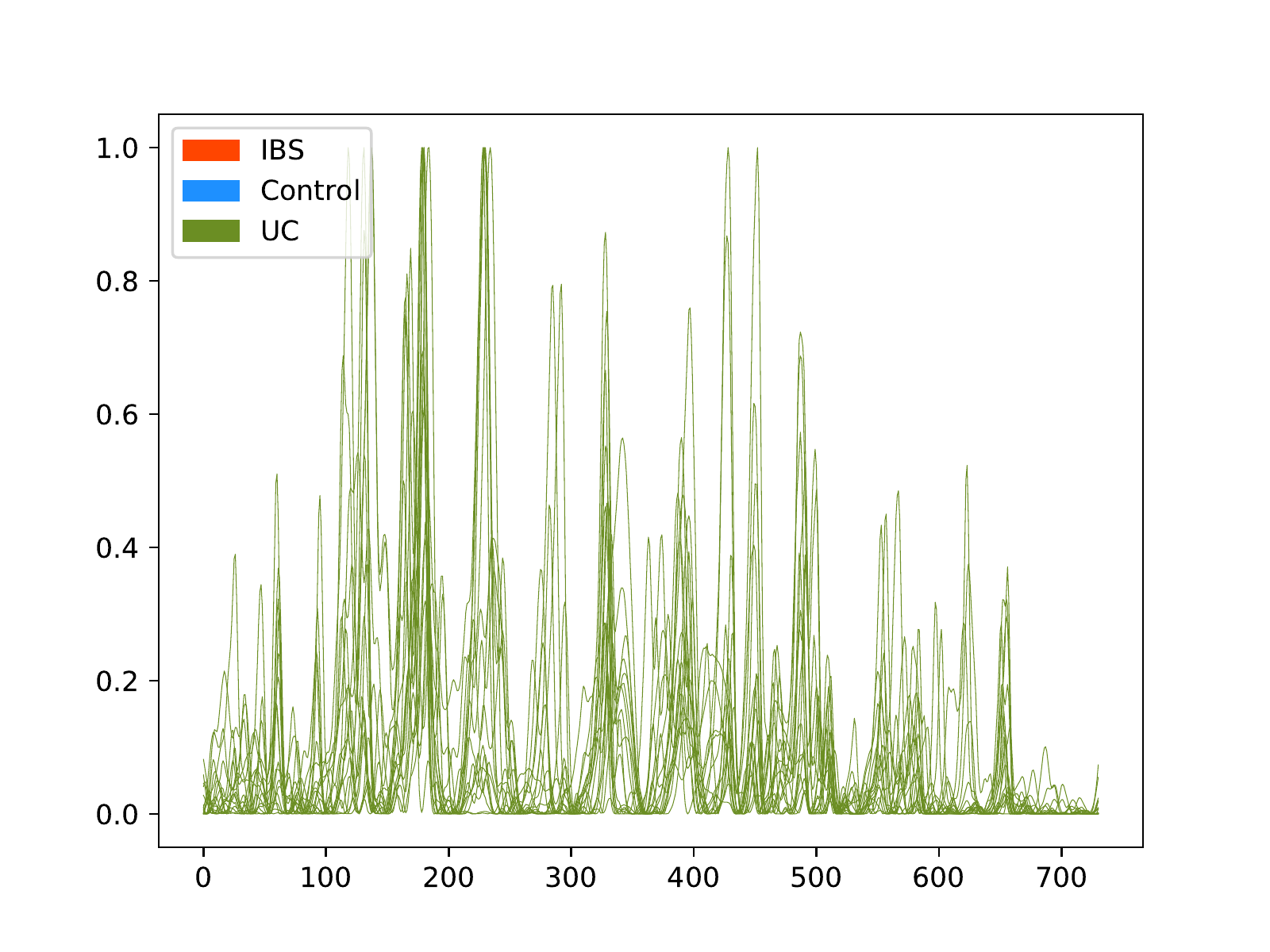}}]
  [\scalebox{0.47}{\includegraphics{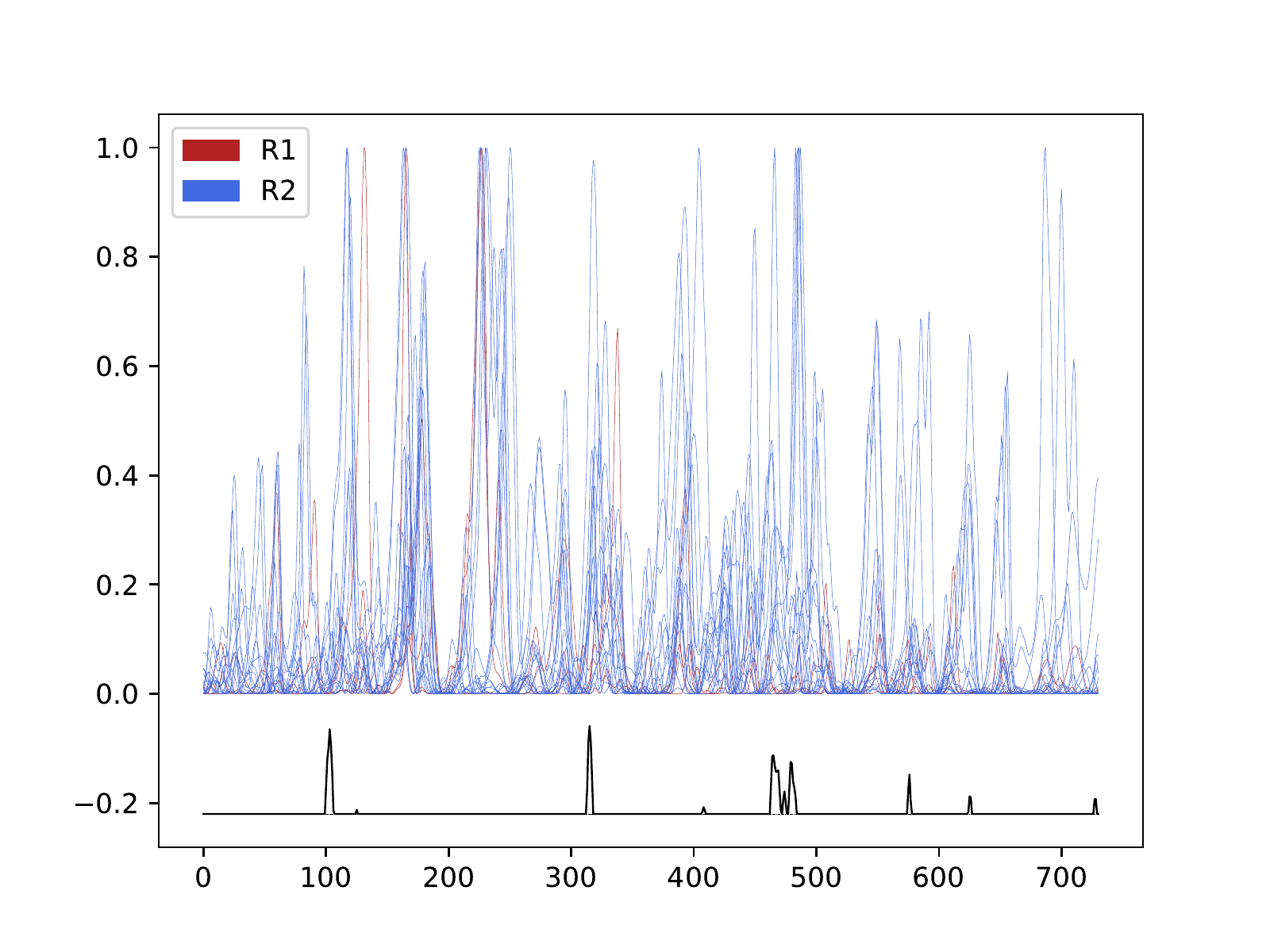}}  \\ $w_{neg} - f_{\mu}$={node[midway,below,font=\scriptsize]{1}}
   [\scalebox{0.47}{\includegraphics{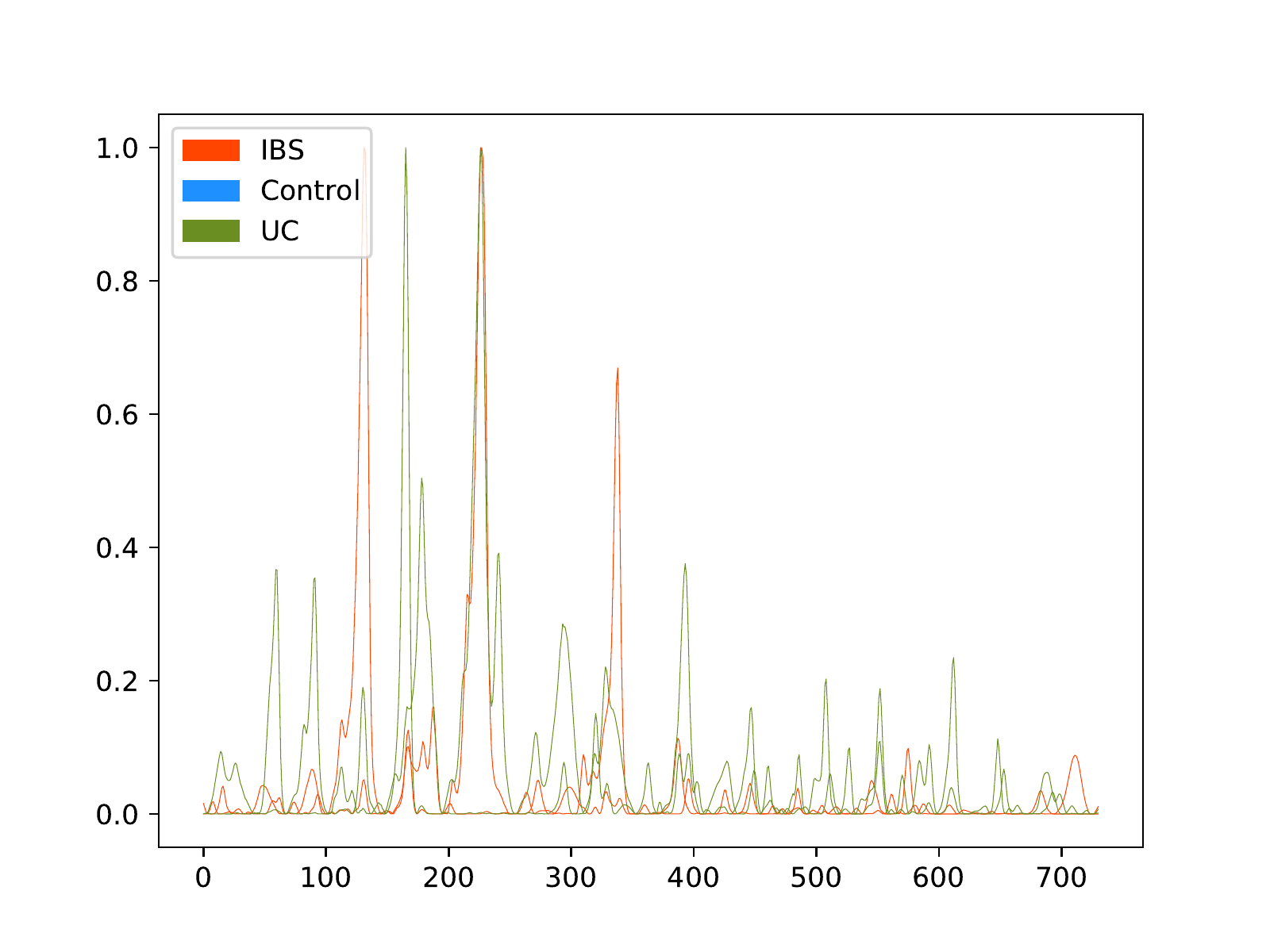}}]
   [\scalebox{0.47}{\includegraphics{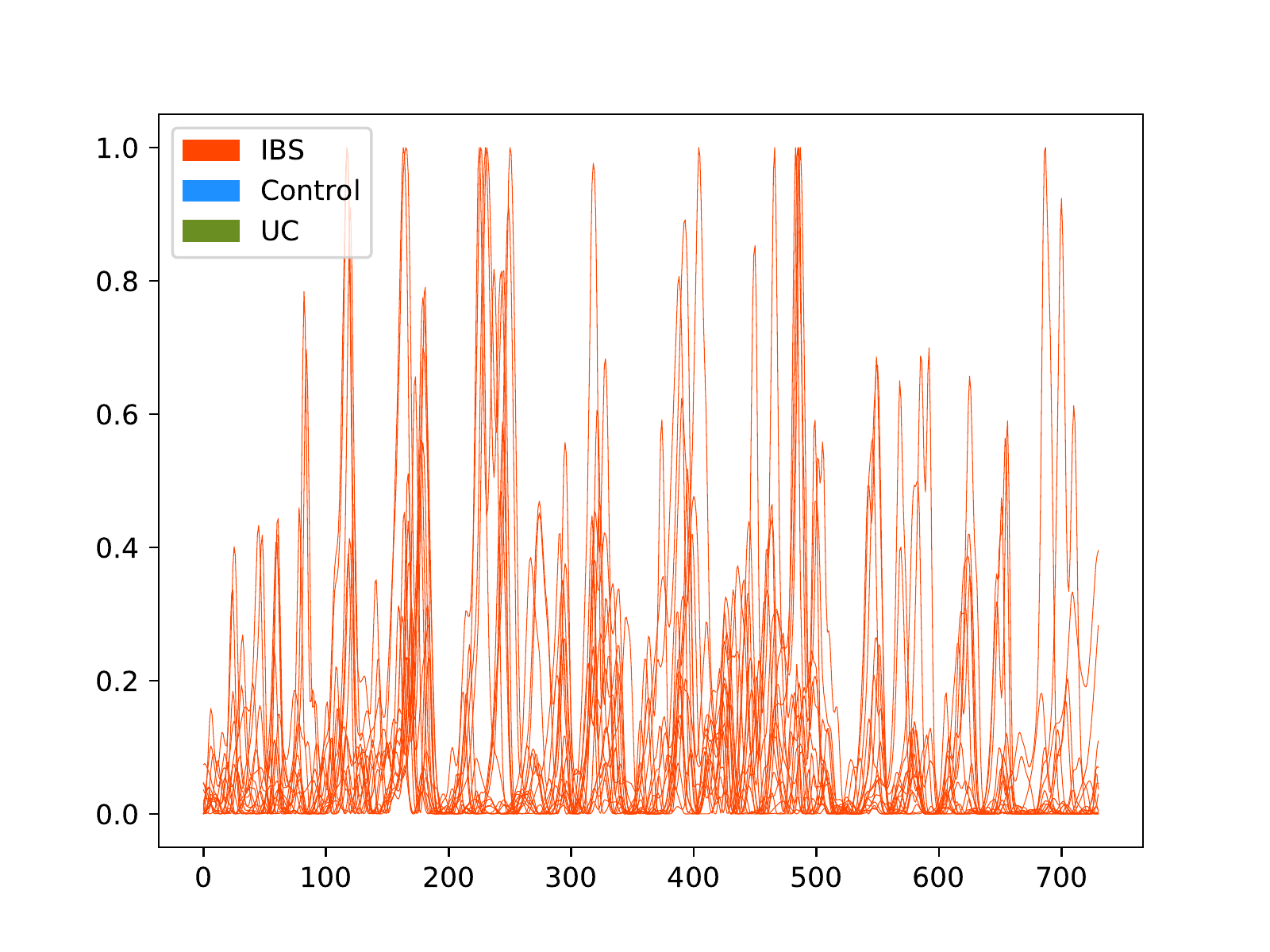}}]
  ]
 ]
]
\end{forest}
}
\caption{Gut Microbiota DGGE: fully grown tree}
\label{fig:dgge_fulltree}
\end{figure}

This second application is a multiclass classification problem that aims to discriminate between subjects with ulcerative colitis (UC), irritable bowel syndrome (IBS) or healthy controls, by analyzing their gut microbiota compositions, where different localized patterns seem to be associated with the different conditions. The input data is obtained through denaturing gradient gel electrophoresis (DGGE) of faecal samples, and the details of the data acquisition and processing are explained in \cite{dgge}. The dataset is shown in Figure \ref{fig:dgge} and is composed of $N$=96 discretized functional samples of length $p$=731, presmoothed and without missing values, available at \url{http://csr.quadram.ac.uk/wp-content/uploads/DGGE_Timepoint_1.zip}. Table \ref{tab:dgge} shows the results, where for the ensemble methods the feature extraction approach is less effective, while locally learning the weights allows $\mu$CART to provide a higher score, especially if we consider the fact that this is a multiclass problem. It is also worth to note that the difference in height between the trees is lower, which could again be a consequence of how we deal with multiple classes while learning the measure in the nodes. The $\mu$CART full tree is depicted in Figure \ref{fig:dgge_fulltree}, where we can observe that the measure inside each decision node becomes more interpretable for deeper nodes. In particular, the deepest internal node selects $w_{neg}$ leading to a sparse weight pattern (as previously noted, we take the absolute value of $w_{neg}$, therefore the baseline is zero and the weights are positive).

\begin{table}[H]
\centering
\caption{\label{tab:dgge} Gut Microbiota DGGE results: classification accuracy and tree height}
\scalebox{1.2}{
\begin{tabular}{l D{,}{\, \pm \,}{-1} D{,}{\, \pm \,}{-1}}
\toprule
\midrule
           
 & \multicolumn{1}{c}{Accuracy \%} & \multicolumn{1}{c}{Tree Height} \\
\midrule
         
\midrule

CART        & 55.3,9  &   3.6,1.3   \\

CART+FE     & 58.5,9  &   5,1.6     \\

RF          & 66,11  &   -         \\
   
RF+FE       & 64.4,10  &   -         \\
  
GBT         & 66,10  &   -         \\

GBT+FE      & 62.2,11   &   -         \\

$\mu$CART   & 71.2,11  &  3.3,0.9     \\
  
\midrule
\bottomrule  
\end{tabular}}
\end{table}

\newpage 

\subsection{Regression}
For the task of regression, we show two spectroscopy applications with the objective to predict the moisture content in either wheat or biscuit dough samples. We again control the growth of all trees by limiting the minimum number of samples required to be at a leaf node, but with mean-square error (mse) instead of Gini index for CART, $\mu$CART and RF, and least squares loss instead of deviance loss for GBT. The number of trees for the ensembles is still fixed to 150, with the same grid search scheme for chosing the maximum number of features for RF. Together with the average tree height, we report the mse on the test set with the corresponding standard deviation, averaged over five random repetitions of 5-fold cross-validation with 3-fold cross-validation for grid search hyperparameter optimization.

\subsubsection{Wheat NIR Spectra}
\begin{figure}[H]
\begin{center}
\includegraphics[scale=0.8]{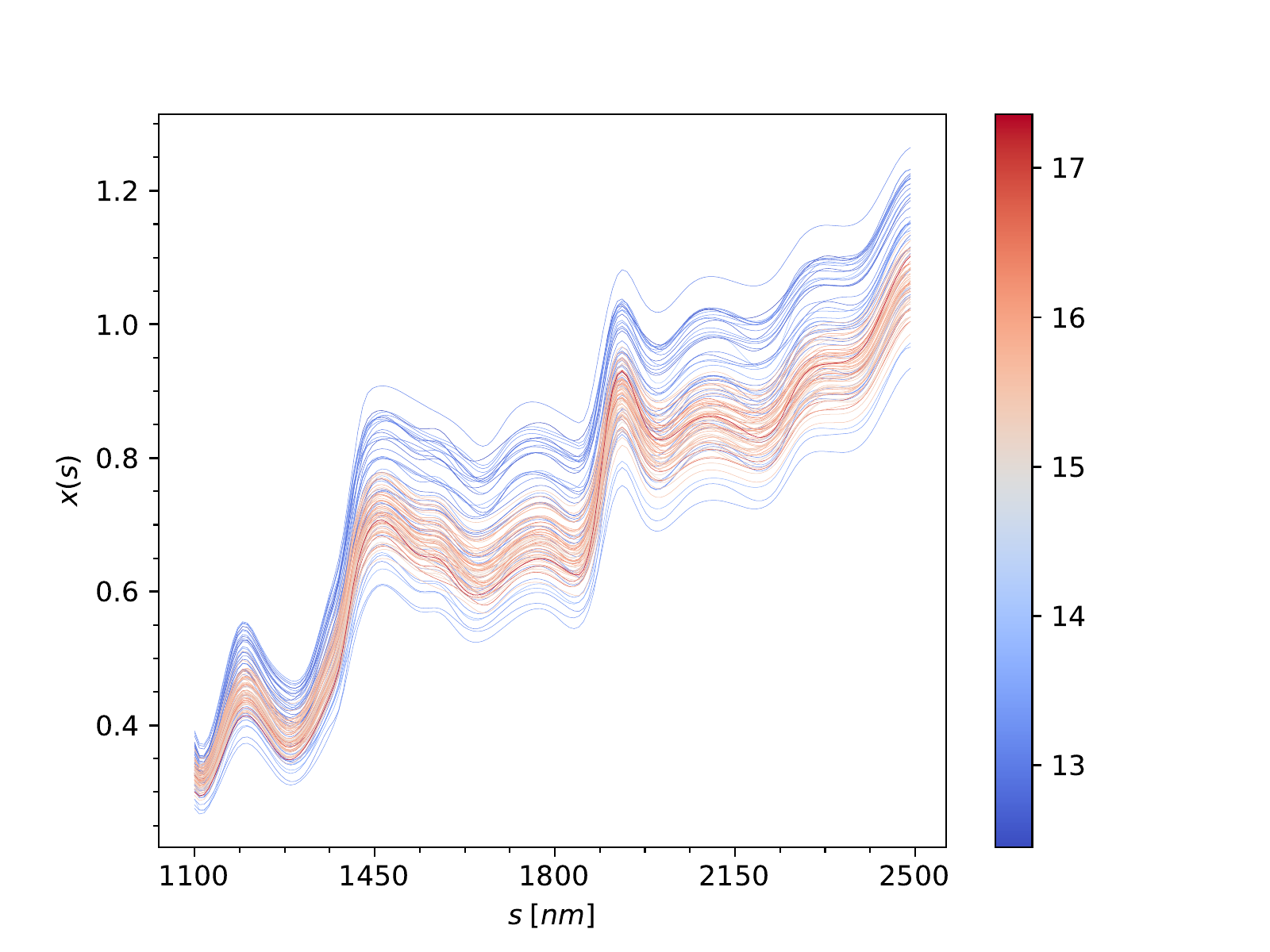}
\caption{\label{fig:wheat} Wheat NIR spectra}
\end{center}
\end{figure}

This dataset is a collection of near-infrared reflectance (NIR) spectra extracted from $N$=100 wheat samples, measured between $1100nm$ and $2500nm$ \citep{wheat}, while the response variable is the moisture content of the sample, with values over the whole dataset of 14.7 $\pm$ 1.37 ($\mu\pm\sigma$). The raw data is available in the CRAN package \textit{fds} \citep{package_fds} and some curves present constant artifacts that may be the result of imputation of missing values. For this reason, we smooth the raw data individually by means of penalized free knots regression splines, with the strength of the penalization that is manually tuned in order to avoid over/under smoothing. This preprocessing is done once before fitting the trees, and therefore all methods are compared on the same data, obtained by evaluating the recovered functions on a grid of length $p$=150, as shown in Figure \ref{fig:wheat}. Table \ref{tab:wheat} shows the results and the tree heights. Like in the previous case study, the chosen features are less effective than directly using all the function evaluations, suggesting that over the whole domain this approach is not favoured for this problem. On the contrary, by extracting those same features on a suitable weighted subset of the domain, $\mu$CART is able to provide the lowest mse at the price of growing more complex trees, as the difference in tree heights between the methods is significant. We omit reporting the fully grown $\mu$CART tree as it would not reasonably fit into a single page.

\begin{table}[H]
\centering
\caption{\label{tab:wheat} Wheat NIR spectra results: regression error (mse) and tree height}
\scalebox{1.2}{
\begin{tabular}{l D{,}{\, \pm \,}{-1} D{,}{\, \pm \,}{-1}}
\toprule
\midrule
           
 & \multicolumn{1}{c}{mse} & \multicolumn{1}{c}{Tree Height} \\
\midrule
         
\midrule

CART        & 1.31,.53  &   4.4,1.9   \\

CART+FE     & 1.57,.54  &   5.4,1.9     \\

RF          & 0.95,.46  &   -         \\
   
RF+FE       & 1.33,.48  &   -         \\
  
GBT         & 0.93,.37  &   -         \\

GBT+FE      & 1.53,.43   &   -         \\

$\mu$CART   & 0.70,.54  &  7.1,0.5     \\
  
\midrule
\bottomrule  
\end{tabular}}
\end{table}

\newpage 

\subsubsection{Biscuit Dough NIR Spectra}
\begin{figure}[H]
\begin{center}
\includegraphics[scale=0.8]{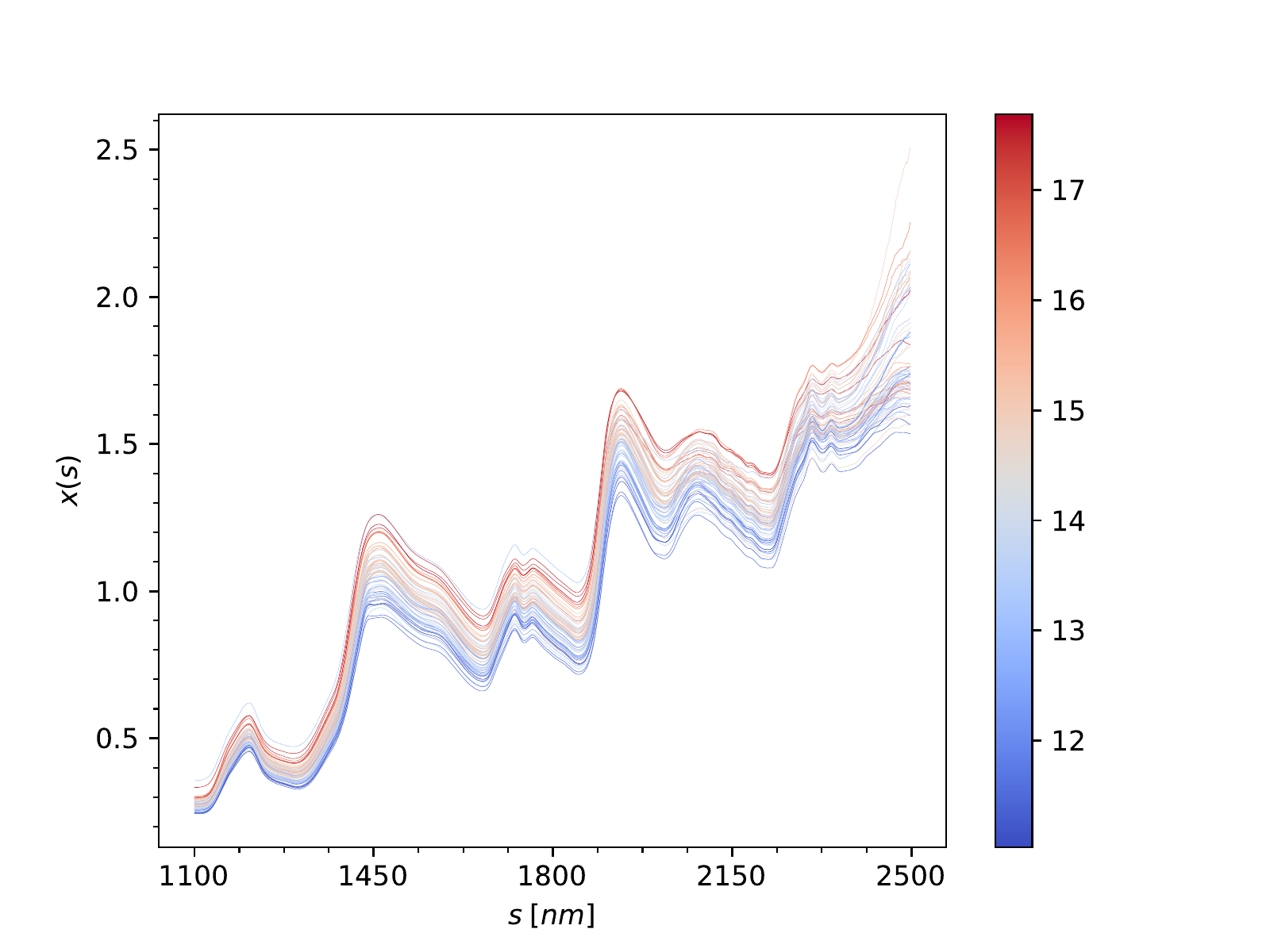}
\caption{\label{fig:dough} Biscuit dough NIR spectra}
\end{center}
\end{figure}

\begin{figure}[H]
\centering
\begin{forest}
[\scalebox{0.07}{\includegraphics[width=6\textwidth]{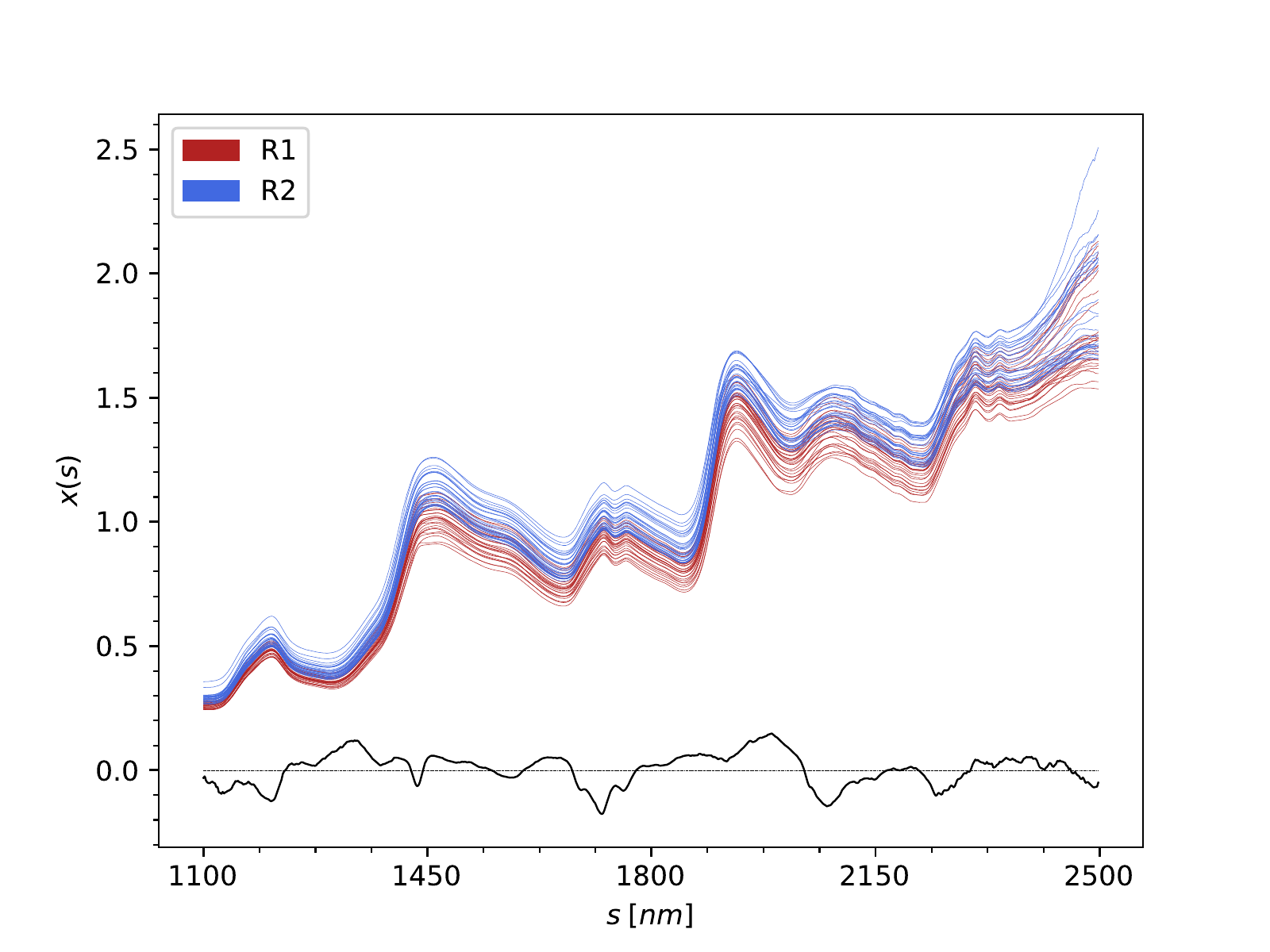}} \\ $w_{sgn} - f_{\mu}$={node[midway,below,font=\scriptsize]{1}}
 [\scalebox{0.06}{\includegraphics[width=5\textwidth]{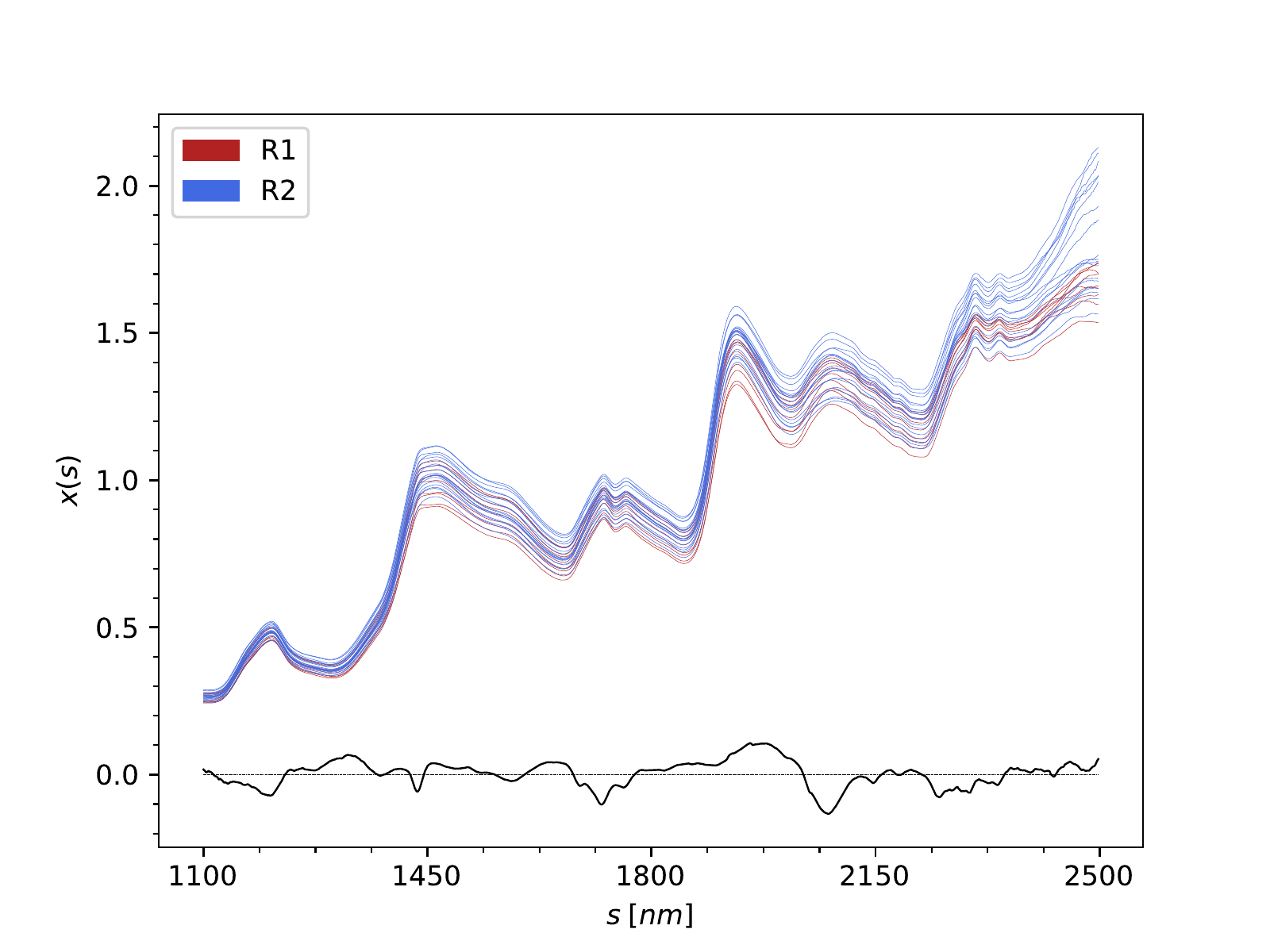}} \\ $w_{sgn} - f_{\sigma^{2}}$={node[midway,below,font=\scriptsize]{1}}
  [\scalebox{0.04}{\includegraphics[width=4.5\textwidth]{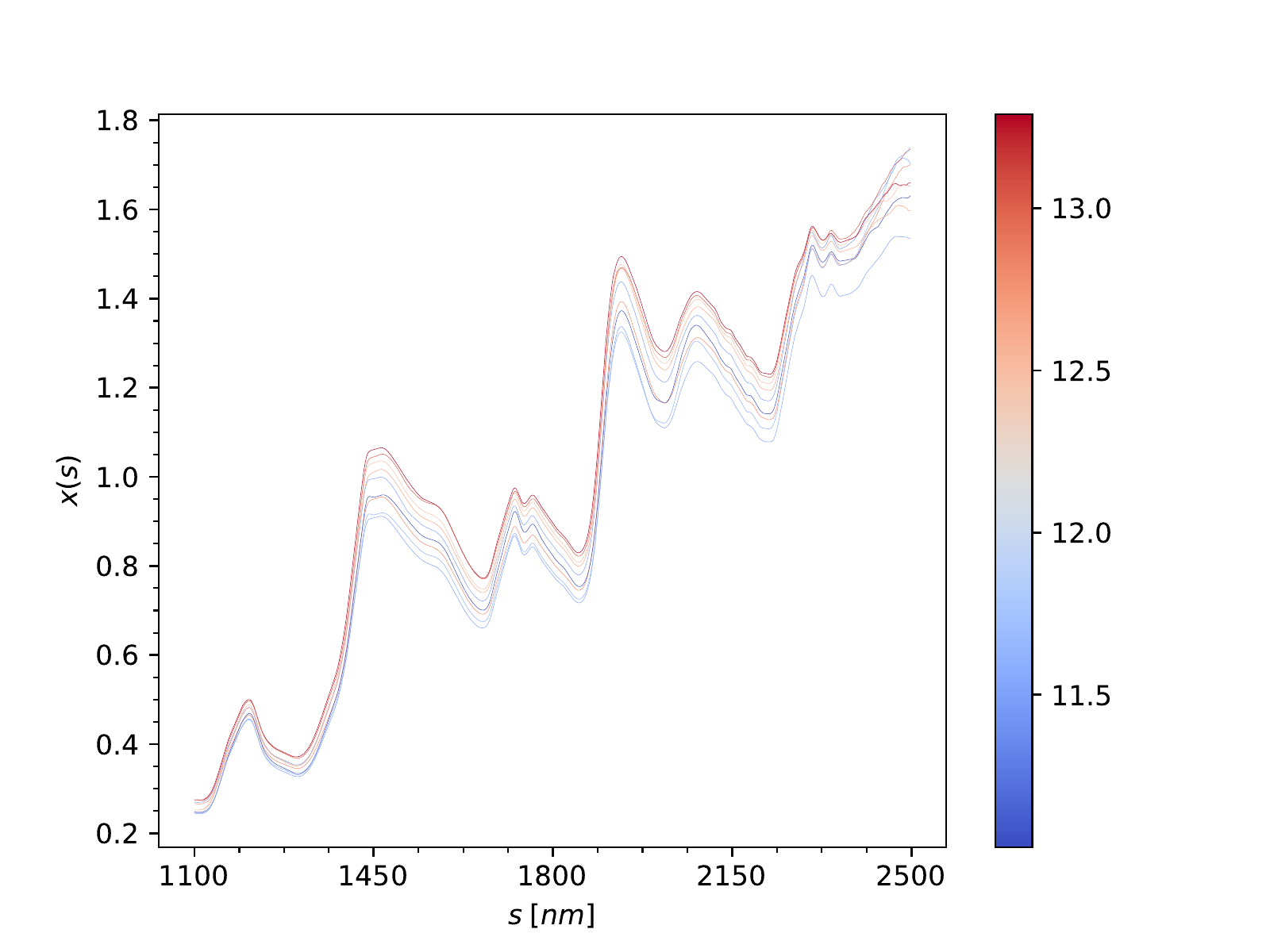}}]
  [\scalebox{0.04}{\includegraphics[width=4.5\textwidth]{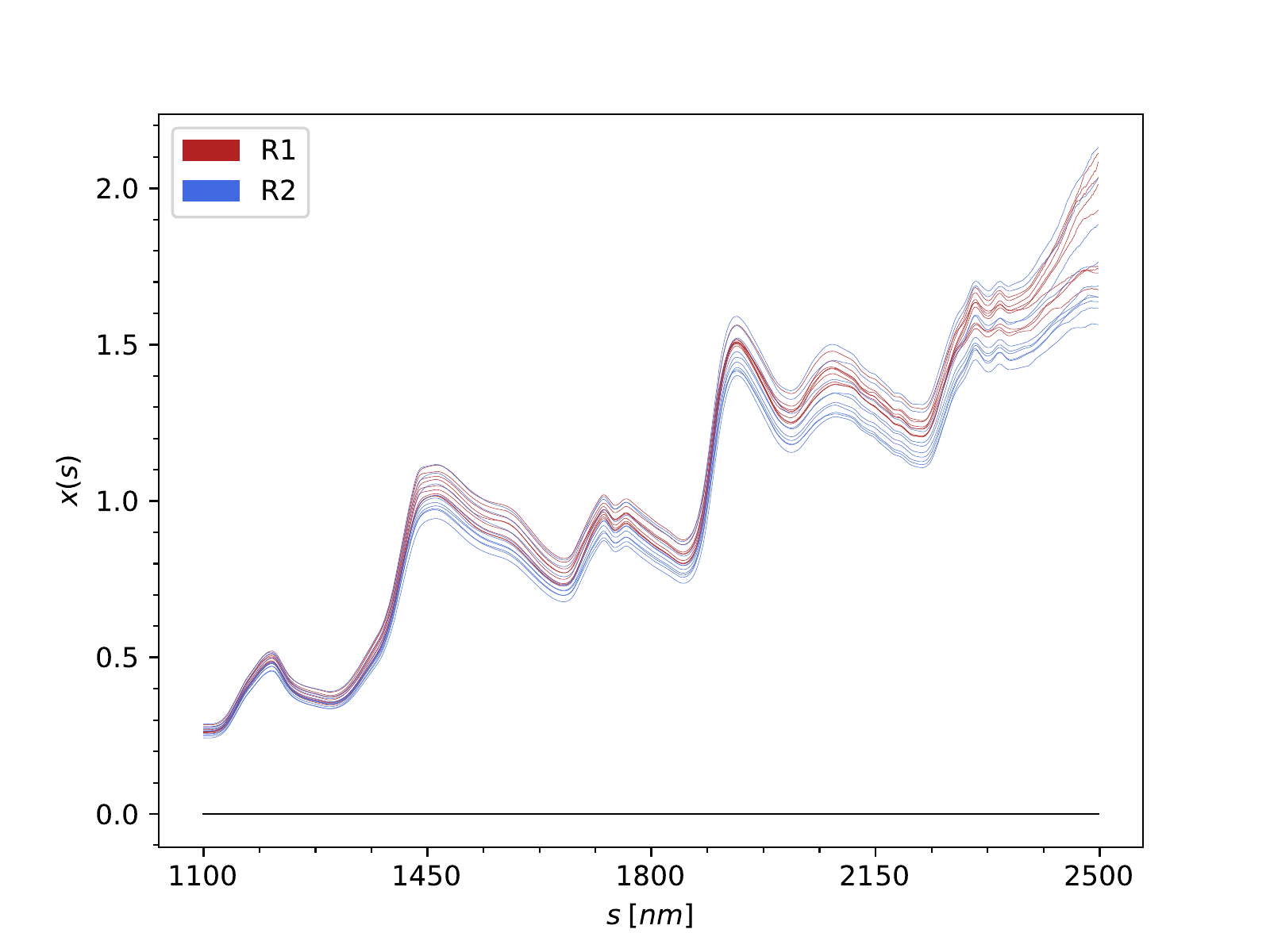}} \\ $w_{uni} - f_{\cos\theta}$={node[midway,below,font=\scriptsize]{1}}
   [\scalebox{0.03}{\includegraphics[width=4.5\textwidth]{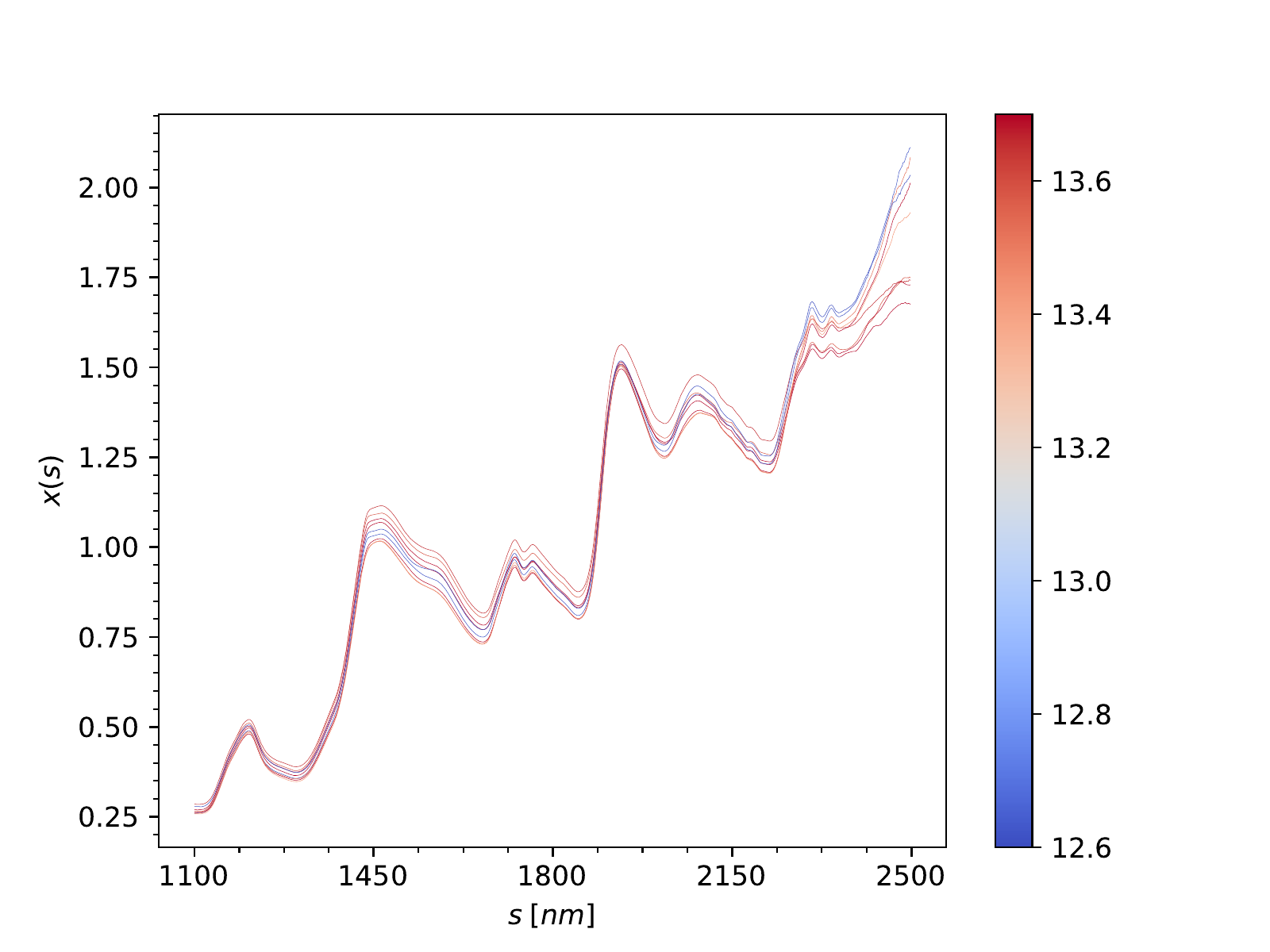}}]
   [\scalebox{0.03}{\includegraphics[width=4.5\textwidth]{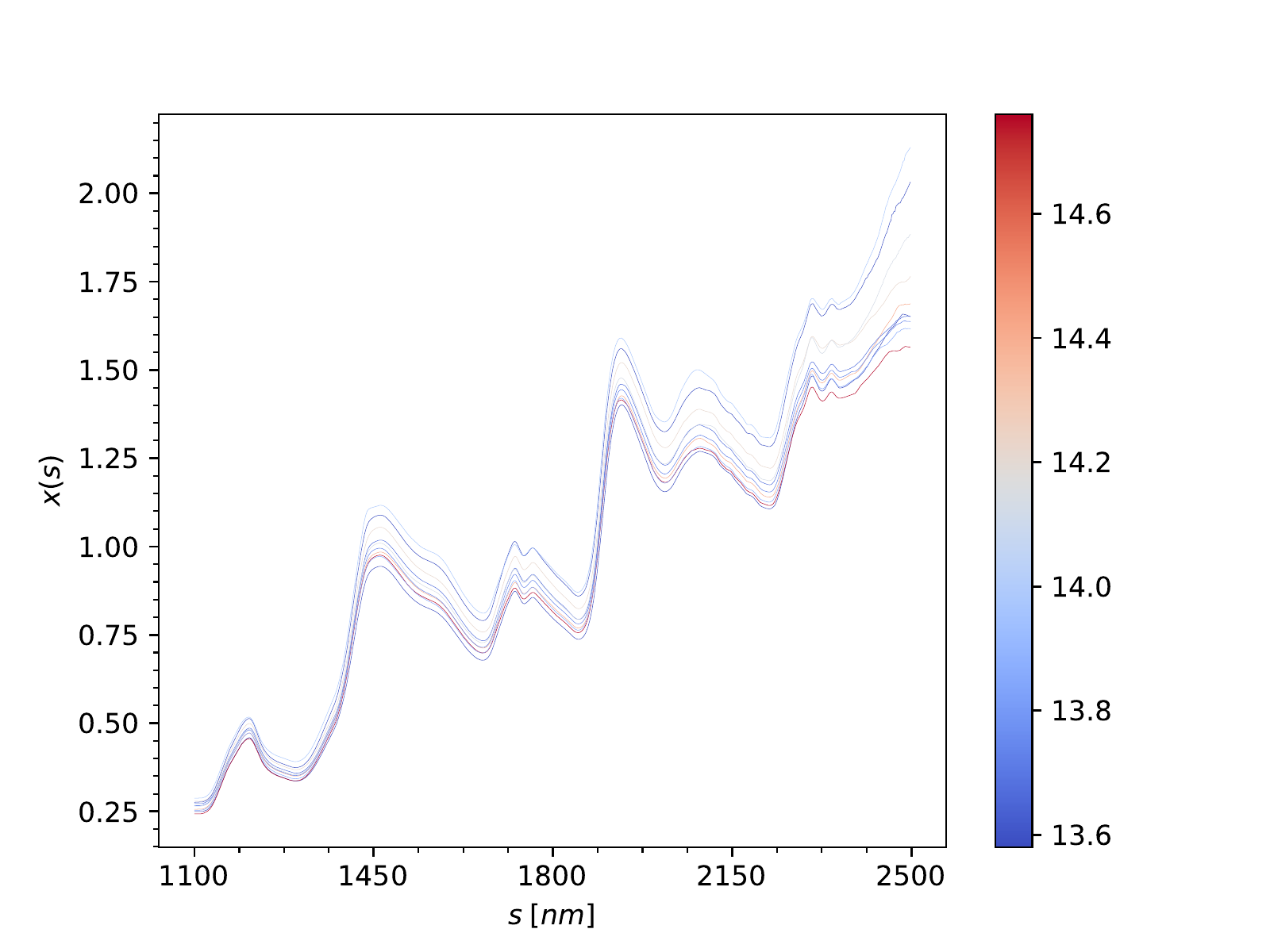}}]
  ]	
 ]
 [\scalebox{0.06}{\includegraphics[width=5\textwidth]{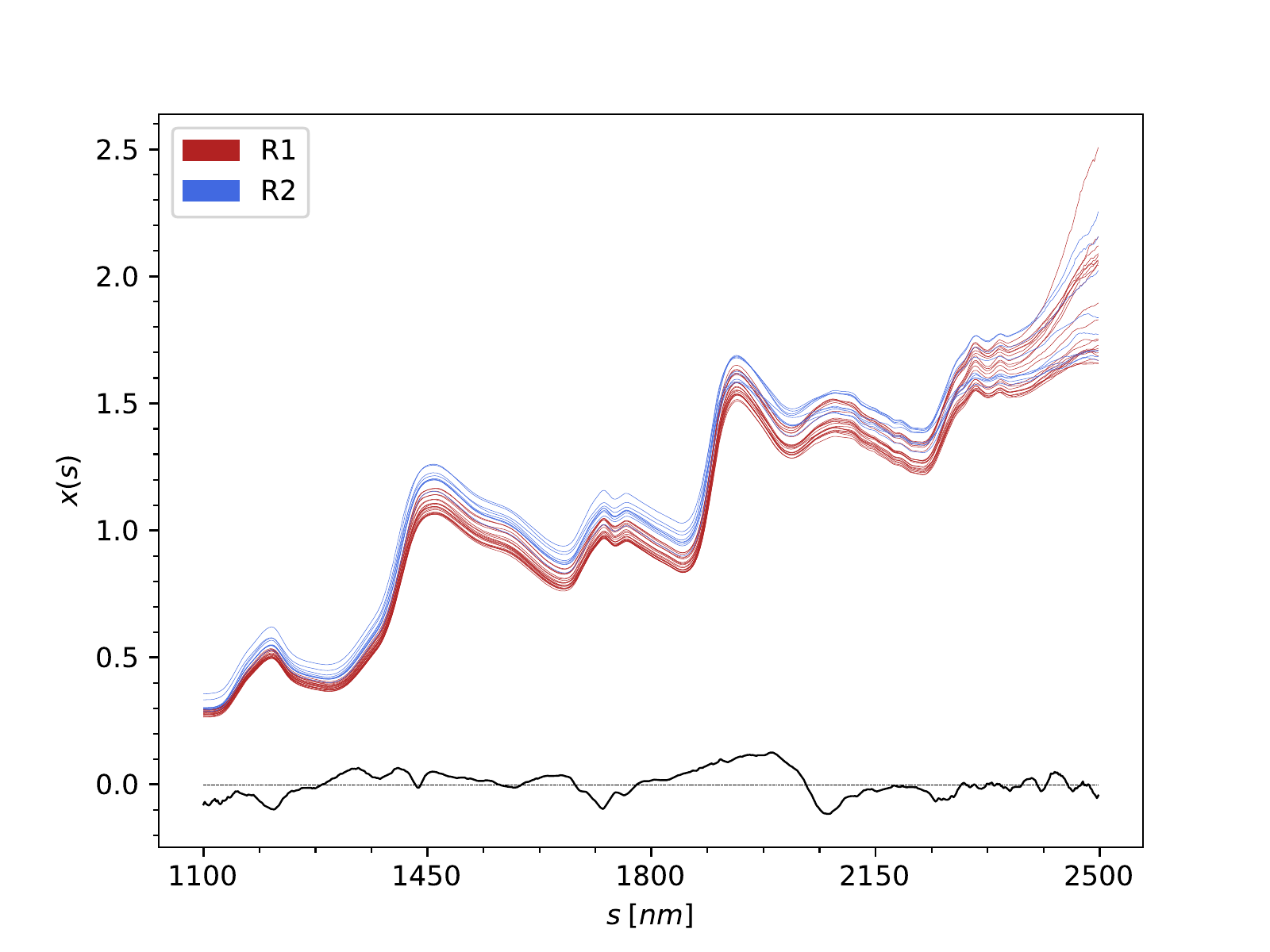}} \\ $w_{sgn} - f_{\mu}$={node[midway,below,font=\scriptsize]{1}}
  [\scalebox{0.04}{\includegraphics[width=4.5\textwidth]{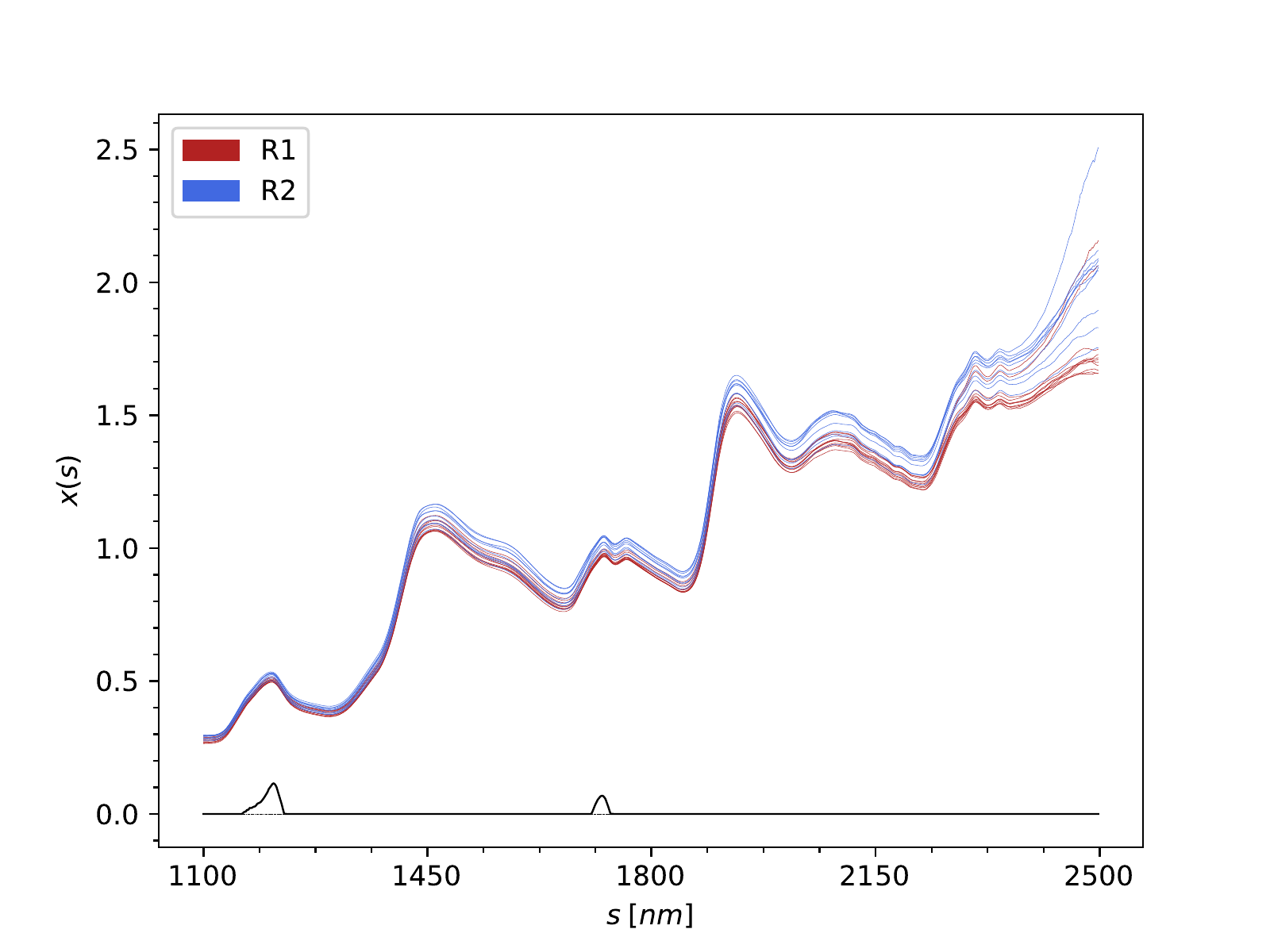}} \\ $w_{neg} - f_{\sigma^{2}}$={node[midway,below,font=\scriptsize]{1}}
   [\scalebox{0.03}{\includegraphics[width=4.5\textwidth]{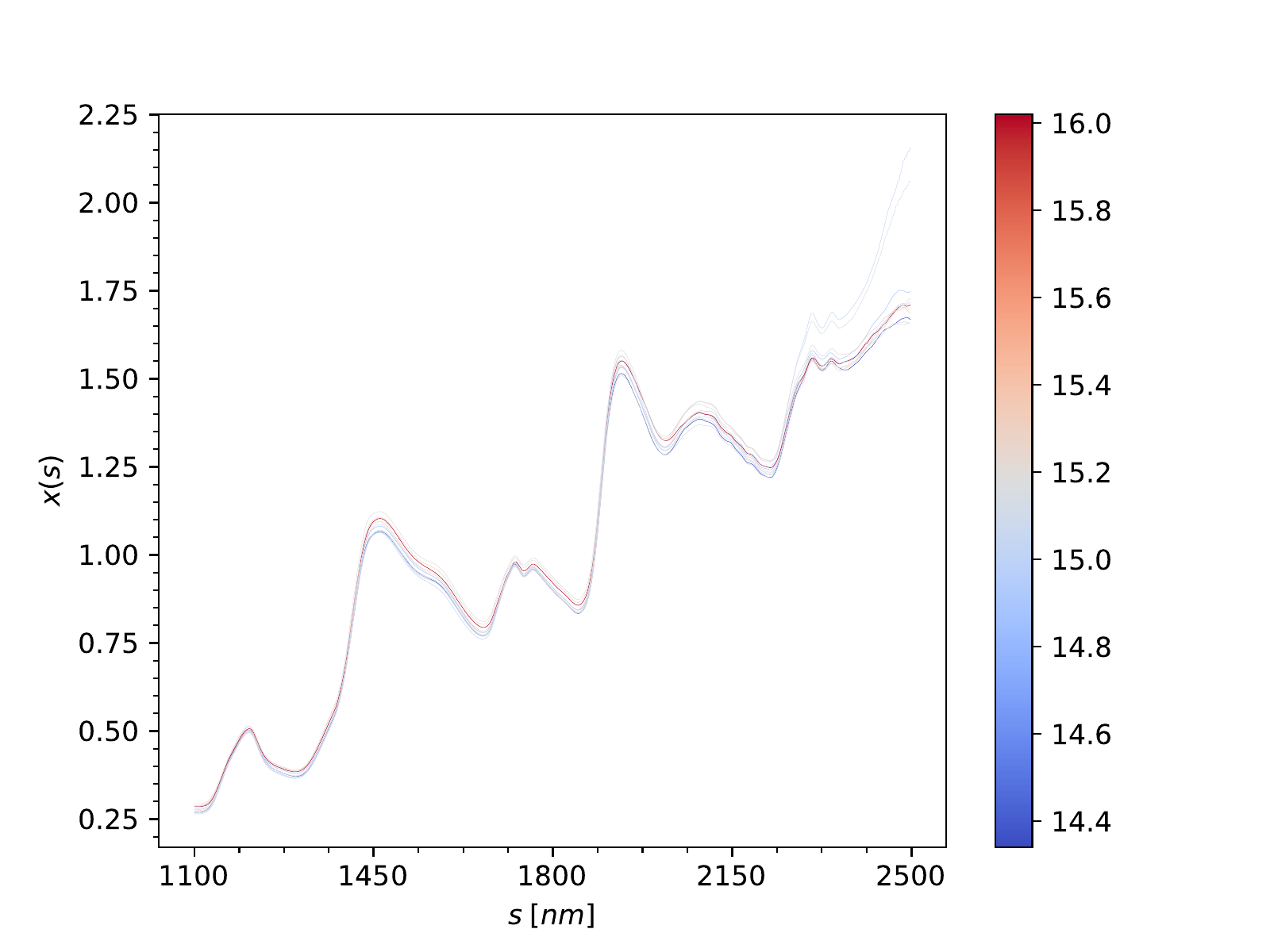}}]
   [\scalebox{0.03}{\includegraphics[width=4.5\textwidth]{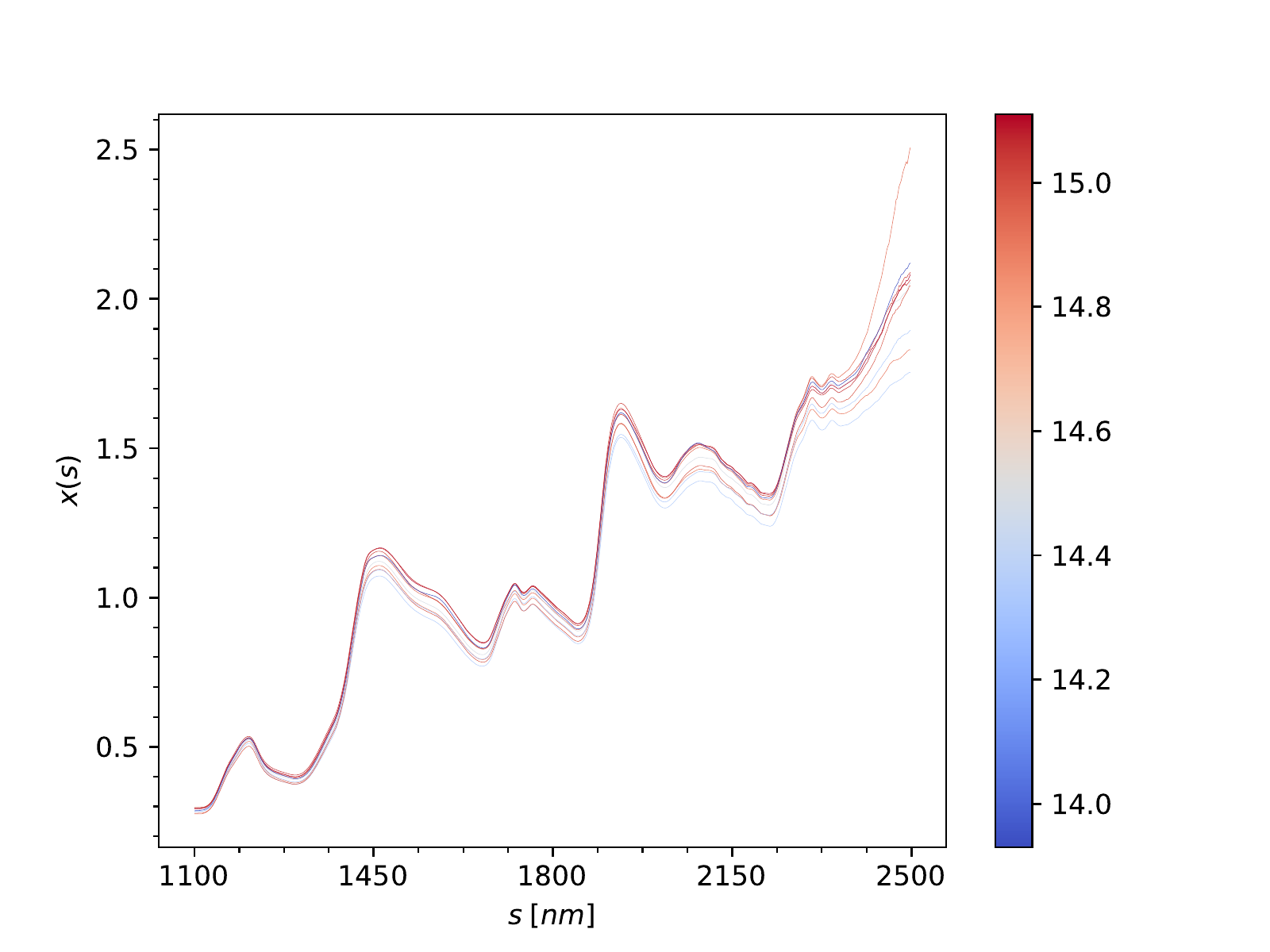}}]
  ]
  [\scalebox{0.04}{\includegraphics[width=4.5\textwidth]{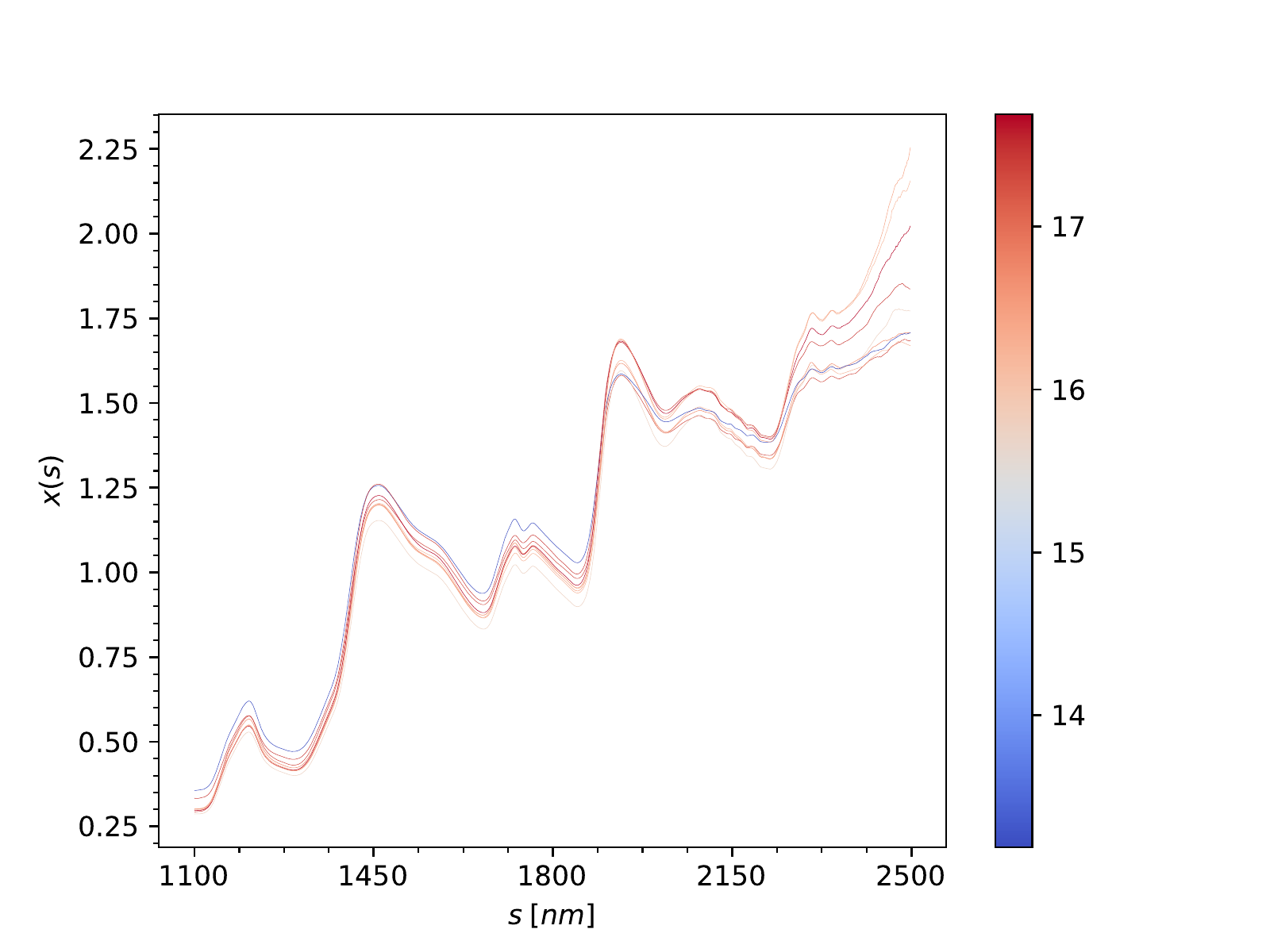}}]
 ]
]
\end{forest}
\caption{Biscuit dough NIR spectra: fully grown tree}
\label{fig:dough_fulltree}
\end{figure}

The second regression application deals with predicting the moisture content of biscuit dough (14.2 $\pm$ 1.47), from NIR spectra measured between $1100nm$ and $2498nm$. The raw data is also available in the CRAN package \textit{fds} \citep{package_fds}, and is comprised of $N$=72 curves evaluated in $p$=700 equispaced wavelengths, without missing values or conspicuous noise, as shown in Figure \ref{fig:dough}. This dataset was originally studied in a Bayesian wavelet framework \citep{dough}, with a different preprocessing in order to reduce compute time, we instead use the evaluations on the whole domain. Table \ref{tab:dough} shows the results with the corresponding tree heights. The relative errors between the methods and preprocessing approaches are similar to the previous case study, suggesting that locally weighted features are an effective approach for moisture prediction in the $1100nm$-$2498nm$ band. The tree heights are instead comparable between the methods, with Figure \ref{fig:dough_fulltree} showing the fitted $\mu$CART tree. As observed in the gut microbiota application, the learned representations become more interpretable with the depth of the nodes, and in particular we find a similarly sparse pattern in the right branch, while the left one selects the uniform weights with the cosine distance as feature extractor.

\begin{table}[H]
\centering
\caption{\label{tab:dough} Biscuit dough NIR spectra results: regression error (mse) and tree height}
\scalebox{1.2}{
\begin{tabular}{l D{,}{\, \pm \,}{-1} D{,}{\, \pm \,}{-1}}
\toprule
\midrule
           
 & \multicolumn{1}{c}{mse} & \multicolumn{1}{c}{Tree Height} \\
\midrule
         
\midrule

CART        & 1.24,.49  &   3.4,1.9   \\

CART+FE     & 1.58,.52  &   3.6,1.4     \\

RF          & 0.90,.41  &   -         \\
   
RF+FE       & 1.32,.42  &   -         \\
  
GBT         & 1.07,.49  &   -         \\

GBT+FE      & 1.52,.49   &   -         \\

$\mu$CART   & 0.74,.38  &  3.8,0.4     \\
  
\midrule
\bottomrule  
\end{tabular}}
\end{table}

\newpage

\newpage 

\section{Conclusion}
\label{conc}
In this work we have proposed a novel nonparametric tree-based tool for classification and regression in the context of functional and high dimensional smooth data. In particular, we introduced a set of splitting rules that are able to leverage the functional data model at the node level, while at the same time maintaining the capability of handling multiple and heterogeneous data sources, as the growing procedure fits within the known CART greedy algorithm. By leveraging the concept of weighted $L^{2}$ spaces, it is possible to learn meaningful representations for each internal node of the tree, extracting locally informative features that allow for an interpretable method with low generalization error and variable selection properties. The input functions are therefore seen as a single unit, for instead of relying on the single evaluations to perform the split, we select between a set of user defined adaptively weighted features. Our splitting rule can be seen as a specific instance of oblique splitting, and given that a direct optimization of the node is intractable, we proposed to learn the split by means of constrained convex optimization with interior point methods, modeling the linear combination as the weight function that induces the measure of the weighted $L^{2}$ space in the current node. The effectiveness of our method has been shown in both classification and regression tasks with one or multiple scalar-valued functional covariates defined on unidimensional domains, through simulation studies and real world spectrometry applications. In particular, when the chosen features are in fact informative for the task at hand, our method can have lower error than ensemble methods like random forest and gradient boosted trees. As our work focused only on simpler functional covariates, further extensions can be developed for more complex data like vector-valued functions defined on multidimensional domains. Other interesting aspects could be bagging and randomized selection of the splitting variable/feature extractor.

\section*{Acknowledgements}
Edoardo Belli was financially supported by the ABB-Politecnico di Milano Joint Research Center through the PhD scholarship \textit{"Development and prototyping of distributed control systems for electric networks based on advanced statistical models for the analysis of complex data"}.

\bibliographystyle{abbrvnat}

\bibliography{template}

\begin{thebibliography}{77}
\providecommand{\natexlab}[1]{#1}
\providecommand{\url}[1]{\texttt{#1}}
\expandafter\ifx\csname urlstyle\endcsname\relax
  \providecommand{\doi}[1]{doi: #1}\else
  \providecommand{\doi}{doi: \begingroup \urlstyle{rm}\Url}\fi

\bibitem[Alonso et~al.(2012)Alonso, Casado, and Romo]{fda_weighted}
A.~M. Alonso, D.~Casado, and J.~Romo.
\newblock Supervised classification for functional data: A weighted distance
  approach.
\newblock \emph{Computational Statistics \& Data Analysis}, 56\penalty0
  (1):\penalty0 2334--2346, 2012.

\bibitem[Balakrishnan and Madigan(2006)]{fda_dtfunctional}
S.~Balakrishnan and D.~Madigan.
\newblock Decision trees for functional variables.
\newblock In \emph{International Conference on Data Mining}, 2006.

\bibitem[Bennett and Mangasarian(1992)]{disclinprog}
K.~P. Bennett and O.~Mangasarian.
\newblock Robust linear programming discrimination of two linearly inseparable
  sets.
\newblock \emph{Optimization Methods and Software}, 1\penalty0 (1):\penalty0
  23--34, 1992.

\bibitem[Bennett et~al.(2000)Bennett, Cristianini, Shawe-Taylor, and
  Wu]{enlarging}
K.~P. Bennett, N.~Cristianini, J.~Shawe-Taylor, and D.~Wu.
\newblock Enlarging the margins in perceptron decision trees.
\newblock \emph{Machine Learning}, 41\penalty0 (3):\penalty0 295--313, 2000.

\bibitem[Bertsimas and Dunn(2017)]{optimaldt}
D.~Bertsimas and J.~Dunn.
\newblock Optimal classification trees.
\newblock \emph{Machine Learning}, 106\penalty0 (7):\penalty0 1039--1082, 2017.

\bibitem[Breiman(1996)]{bagging}
L.~Breiman.
\newblock Bagging predictors.
\newblock \emph{Machine Learning}, 24\penalty0 (2):\penalty0 123--140, 1996.

\bibitem[Breiman(2001)]{randomforest}
L.~Breiman.
\newblock Random forests.
\newblock \emph{Machine Learning}, 45\penalty0 (1):\penalty0 5--32, 2001.

\bibitem[Breiman et~al.(1984)Breiman, Friedman, Stone, and Olshen]{book_cart}
L.~Breiman, J.~Friedman, C.~J. Stone, and R.~A. Olshen.
\newblock \emph{Classification and regression trees}.
\newblock Wadsworth and Brooks, Belmont, CA, USA, 1984.

\bibitem[Briandet et~al.(1996)Briandet, Kemsley, and Wilson]{coffee}
R.~Briandet, E.~K. Kemsley, and R.~H. Wilson.
\newblock Discrimination of arabica and robusta in instant coffee by fourier
  transform infrared spectroscopy and chemometrics.
\newblock \emph{Journal of Agricultural and Food Chemistry}, 44\penalty0
  (1):\penalty0 170--174, 1996.

\bibitem[Brodley and Utgoff(1995)]{multivdt}
C.~E. Brodley and P.~E. Utgoff.
\newblock Multivariate decision trees.
\newblock \emph{Machine Learning}, 19\penalty0 (1):\penalty0 45--77, 1995.

\bibitem[Brown et~al.(2001)Brown, Fearn, and Vannucci]{dough}
P.~J. Brown, T.~Fearn, and M.~Vannucci.
\newblock Bayesian wavelet regression on curves with application to a
  spectroscopic calibration problem.
\newblock \emph{Journal of the American Statistical Association}, 96\penalty0
  (454):\penalty0 398--408, 2001.

\bibitem[Bul{\`o} and Kontschieder(2014)]{neuraldecforest}
S.~R. Bul{\`o} and P.~Kontschieder.
\newblock Neural decision forests for semantic image labelling.
\newblock In \emph{IEEE Conference on Computer Vision and Pattern Recognition},
  2014.

\bibitem[Cant{\'u}-Paz and Kamath(2003)]{obtreeevo}
E.~Cant{\'u}-Paz and C.~Kamath.
\newblock Inducing oblique decision trees with evolutionary algorithms.
\newblock \emph{IEEE Transactions on Evolutionary Computation}, 7\penalty0
  (1):\penalty0 54--68, 2003.

\bibitem[Cardot et~al.(2003)Cardot, Ferraty, and Sarda]{fda_splineflm}
H.~Cardot, F.~Ferraty, and P.~Sarda.
\newblock Spline estimators for the functional linear model.
\newblock \emph{Statistica Sinica}, 13\penalty0 (3):\penalty0 571--591, 2003.

\bibitem[Cardot et~al.(2007)Cardot, Crambes, Kneip, and
  Sarda]{fda_splineerrors}
H.~Cardot, C.~Crambes, A.~Kneip, and P.~Sarda.
\newblock Smoothing splines estimators in functional linear regression with
  errors-in-variables.
\newblock \emph{Computational Statistics \& Data Analysis}, 51\penalty0
  (10):\penalty0 4832--4848, 2007.

\bibitem[Caruana(1997)]{multitask}
R.~Caruana.
\newblock Multitask learning.
\newblock \emph{Machine Learning}, 28\penalty0 (1):\penalty0 41--75, 1997.

\bibitem[Chaudhuri and Loh(2002)]{quantregtree}
P.~Chaudhuri and W.-Y. Loh.
\newblock Nonparametric estimation of conditional quantiles using quantile
  regression trees.
\newblock \emph{Bernoulli}, 8\penalty0 (5):\penalty0 561--576, 2002.

\bibitem[Cousins and Riondato(2019)]{cdetree}
C.~Cousins and M.~Riondato.
\newblock Ca{D}{E}{T}: interpretable parametric conditional density estimation
  with decision trees and forests.
\newblock \emph{Machine Learning}, 108\penalty0 (8-9):\penalty0 1613--1634,
  2019.

\bibitem[Crambes et~al.(2009)Crambes, Kneip, and Sarda]{fda_smoothsplines}
C.~Crambes, A.~Kneip, and P.~Sarda.
\newblock Smoothing splines estimators for functional linear regression.
\newblock \emph{The Annals of Statistics}, 37\penalty0 (1):\penalty0 35--72,
  2009.

\bibitem[Douzal-Chouakria and Amblard(2012)]{dttimes}
A.~Douzal-Chouakria and C.~Amblard.
\newblock Classification trees for time series.
\newblock \emph{Pattern Recognition}, 45\penalty0 (3):\penalty0 1076--1091,
  2012.

\bibitem[Epifanio(2008)]{fda_features}
I.~Epifanio.
\newblock Shape descriptors for classification of functional data.
\newblock \emph{Technometrics}, 50\penalty0 (3):\penalty0 284--294, 2008.

\bibitem[Ferraty and Vieu(2006)]{book_nonparafda}
F.~Ferraty and P.~Vieu.
\newblock \emph{Nonparametric Functional Data Analysis: Theory and Practice}.
\newblock Springer, New York, NY, USA, 2006.

\bibitem[Friedman(2001)]{boosting}
J.~H. Friedman.
\newblock Greedy function approximation: A gradient boosting machine.
\newblock \emph{The Annals of Statistics}, 29\penalty0 (5):\penalty0
  1189--1232, 2001.

\bibitem[F{\"u}rnkranz et~al.(2020)F{\"u}rnkranz, Kliegr, and
  Paulheim]{ruleplausibility}
J.~F{\"u}rnkranz, T.~Kliegr, and H.~Paulheim.
\newblock On cognitive preferences and the plausibility of rule-based models.
\newblock \emph{Machine Learning}, 109\penalty0 (4):\penalty0 853--898, 2020.

\bibitem[Galeano et~al.(2015)Galeano, Joseph, and Lillo]{fda_maha}
P.~Galeano, E.~Joseph, and R.~E. Lillo.
\newblock The mahalanobis distance for functional data with applications to
  classification.
\newblock \emph{Technometrics}, 57\penalty0 (2):\penalty0 281--291, 2015.

\bibitem[Gama(1999)]{discritrees}
J.~Gama.
\newblock Discriminant trees.
\newblock In \emph{International Conference on Machine Learning}, 1999.

\bibitem[Gama(2004)]{functionaltrees}
J.~Gama.
\newblock Functional trees.
\newblock \emph{Machine Learning}, 55\penalty0 (3):\penalty0 219--250, 2004.

\bibitem[Gondzio(2012)]{iposurvey}
J.~Gondzio.
\newblock Interior point methods 25 years later.
\newblock \emph{European Journal of Operational Research}, 218\penalty0
  (3):\penalty0 587--601, 2012.

\bibitem[Hart et~al.(2017)Hart, Laird, Watson, Woodruff, Hackebeil, Nicholson,
  and Siirola]{book_pyomo}
W.~E. Hart, C.~D. Laird, J.-P. Watson, D.~L. Woodruff, G.~A. Hackebeil, B.~L.
  Nicholson, and J.~D. Siirola.
\newblock \emph{Pyomo--optimization modeling in python}.
\newblock Springer Science \& Business Media, New York, NY, USA, 2017.

\bibitem[Hastie et~al.(2009)Hastie, Tibshirani, and Friedman]{book_esl}
T.~Hastie, R.~Tibshirani, and J.~Friedman.
\newblock \emph{The Elements of Statistical Learning}.
\newblock Springer, New York, NY, USA, 2009.

\bibitem[Heath et~al.(1993)Heath, Kasif, and Salzberg]{sadt}
D.~Heath, S.~Kasif, and S.~Salzberg.
\newblock Induction of oblique decision trees.
\newblock In \emph{International Joint Conference on Artificial Intelligence},
  1993.

\bibitem[Horv\'{a}th and Kokoszka(2012)]{book_inferfda}
L.~Horv\'{a}th and P.~Kokoszka.
\newblock \emph{Inference for functional data with applications}.
\newblock Springer, New York, NY, USA, 2012.

\bibitem[Hothorn et~al.(2006)Hothorn, Hornik, and Zeileis]{unbrecurpart}
T.~Hothorn, K.~Hornik, and A.~Zeileis.
\newblock Unbiased recursive partitioning: A conditional inference framework.
\newblock \emph{Journal of Computational and Graphical Statistics}, 15\penalty0
  (3):\penalty0 651--674, 2006.

\bibitem[James et~al.(2000)James, Hastie, and Sugar]{fda_pcasparse}
G.~M. James, T.~J. Hastie, and C.~A. Sugar.
\newblock Principal component models for sparse functional data.
\newblock \emph{Biometrika}, 87\penalty0 (3):\penalty0 587--602, 2000.

\bibitem[Jordan and Jacobs(1994)]{hme}
M.~I. Jordan and R.~A. Jacobs.
\newblock Hierarchical mixtures of experts and the em algorithm.
\newblock \emph{Neural Computation}, 6\penalty0 (2):\penalty0 181--214, 1994.

\bibitem[Kalivas(1997)]{wheat}
J.~H. Kalivas.
\newblock Two data sets of near infrared spectra.
\newblock \emph{Chemometrics and Intelligent Laboratory Systems}, 37\penalty0
  (2):\penalty0 255--259, 1997.

\bibitem[Kim and Loh(2001)]{cruise}
H.~Kim and W.-Y. Loh.
\newblock Classification trees with unbiased multiway splits.
\newblock \emph{Journal of the American Statistical Association}, 96\penalty0
  (454):\penalty0 589--604, 2001.

\bibitem[Kontschieder et~al.(2015)Kontschieder, Fiterau, Criminisi, and
  Bul{\`o}]{deepneuraldecforest}
P.~Kontschieder, M.~Fiterau, A.~Criminisi, and S.~R. Bul{\`o}.
\newblock Deep neural decision forests.
\newblock In \emph{IEEE International Conference on Computer Vision}, 2015.

\bibitem[Landwehr et~al.(2005)Landwehr, Hall, and Frank]{logisticmodeltrees}
N.~Landwehr, M.~Hall, and E.~Frank.
\newblock Logistic model trees.
\newblock \emph{Machine Learning}, 59\penalty0 (1-2):\penalty0 161--205, 2005.

\bibitem[Lee and Shih(2006)]{ctreemultilabel2}
T.-H. Lee and Y.-S. Shih.
\newblock Unbiased variable selection for classification trees with
  multivariate responses.
\newblock \emph{Computational Statistics \& Data Analysis}, 51\penalty0
  (2):\penalty0 659--667, 2006.

\bibitem[Li and Yu(2008)]{fda_segm}
B.~Li and Q.~Yu.
\newblock Classification of functional data: A segmentation approach.
\newblock \emph{Computational Statistics \& Data Analysis}, 52\penalty0
  (1):\penalty0 4790--4800, 2008.

\bibitem[Liu and Tsang(2017)]{dtultrahigh}
W.~Liu and I.~W. Tsang.
\newblock Making decision trees feasible in ultrahigh feature and label
  dimensions.
\newblock \emph{Journal of Machine Learning Research}, 18\penalty0
  (81):\penalty0 1--36, 2017.

\bibitem[Loh(2002)]{guide_reg}
W.-Y. Loh.
\newblock Regression trees with unbiased variable selection and interaction
  detection.
\newblock \emph{Statistica Sinica}, 12\penalty0 (2):\penalty0 361--386, 2002.

\bibitem[Loh(2009)]{guide_class}
W.-Y. Loh.
\newblock Improving the precision of classification trees.
\newblock \emph{Annals of Applied Statistics}, 3\penalty0 (4):\penalty0
  1710--1737, 2009.

\bibitem[Loh and Shih(1997)]{quest}
W.-Y. Loh and Y.-S. Shih.
\newblock Split selection methods for classification trees.
\newblock \emph{Statistica Sinica}, 7\penalty0 (4):\penalty0 815--840, 1997.

\bibitem[Loh and Vanichsetakul(1988)]{discritreesclass}
W.-Y. Loh and N.~Vanichsetakul.
\newblock Tree-structured classification via generalized discriminant analysis.
\newblock \emph{Journal of the American Statistical Association}, 83\penalty0
  (403):\penalty0 715--725, 1988.

\bibitem[Loh and Zheng(2013)]{guidemultivrespo}
W.-Y. Loh and W.~Zheng.
\newblock Regression trees for longitudinal and multiresponse data.
\newblock \emph{The Annals of Applied Statistics}, 7\penalty0 (1):\penalty0
  495--522, 2013.

\bibitem[Martin(1997)]{pmetrictreestopping}
J.~K. Martin.
\newblock An exact probability metric for decision tree splitting and stopping.
\newblock \emph{Machine Learning}, 28\penalty0 (2-3):\penalty0 257--291, 1997.

\bibitem[Meinshausen(2006)]{quantregforest}
N.~Meinshausen.
\newblock Quantile regression forests.
\newblock \emph{Journal of Machine Learning Research}, 7\penalty0
  (Jun):\penalty0 983--999, 2006.

\bibitem[Meinshausen(2013)]{signconstr}
N.~Meinshausen.
\newblock Sign-constrained least squares estimation for high-dimensional
  regression.
\newblock \emph{Electronic Journal of Statistics}, 7\penalty0 (1):\penalty0
  1607--1631, 2013.

\bibitem[M{\"o}ller and Gertheiss(2018)]{fda_classtfunctional}
A.~M{\"o}ller and J.~Gertheiss.
\newblock A classification tree for functional data.
\newblock In \emph{International Workshop on Statistical Modeling}, 2018.

\bibitem[M{\"o}ller et~al.(2016)M{\"o}ller, Tutz, and Gertheiss]{fda_rf}
A.~M{\"o}ller, G.~Tutz, and J.~Gertheiss.
\newblock Random forests for functional covariates.
\newblock \emph{Journal of Chemometrics}, 30\penalty0 (1):\penalty0 715--725,
  2016.

\bibitem[Murthy et~al.(1994)Murthy, Kasif, and Salzberg]{oc1}
S.~K. Murthy, S.~Kasif, and S.~Salzberg.
\newblock A system for induction of oblique decision trees.
\newblock \emph{Journal of Artificial Intelligence Research}, 2\penalty0
  (1):\penalty0 1--32, 1994.

\bibitem[Nerini and Ghattas(2007)]{fda_treeresponse_ocean}
D.~Nerini and B.~Ghattas.
\newblock Classifying densities using functional regression trees: Applications
  in oceanology.
\newblock \emph{Computational Statistics \& Data Analysis}, 51\penalty0
  (10):\penalty0 4984--4993, 2007.

\bibitem[Noh et~al.(2004)Noh, Song, and Park]{ctreemultilabel}
H.~G. Noh, M.~S. Song, and S.~H. Park.
\newblock An unbiased method for constructing multilabel classification trees.
\newblock \emph{Computational Statistics \& Data Analysis}, 47\penalty0
  (1):\penalty0 149--164, 2004.

\bibitem[Noor et~al.(2010)Noor, Ridgway, Scovell, Kemsley, Lund, Jamieson,
  Johnson, and Narbad]{dgge}
S.~O. Noor, K.~Ridgway, L.~Scovell, E.~K. Kemsley, E.~K. Lund, C.~Jamieson,
  I.~T. Johnson, and A.~Narbad.
\newblock Ulcerative colitis and irritable bowel patients exhibit distinct
  abnormalities of the gut microbiota.
\newblock \emph{BMC Gastroenterology}, 10\penalty0 (134), 2010.

\bibitem[Norouzi et~al.(2015)Norouzi, Collins, Johnson, Fleet, and
  Kohli]{effnongreedy}
M.~Norouzi, M.~D. Collins, M.~Johnson, D.~J. Fleet, and P.~Kohli.
\newblock Efficient non-greedy optimization of decision trees.
\newblock In \emph{Advances in Neural Information Processing Systems}, 2015.

\bibitem[Pande et~al.(2017)Pande, Li, Rajeswaran, Ehrlinger, Kogalur,
  Blackstone, and Ishwaran]{boostdtlongdata}
A.~Pande, L.~Li, J.~Rajeswaran, J.~Ehrlinger, U.~B. Kogalur, E.~H. Blackstone,
  and H.~Ishwaran.
\newblock Boosted multivariate trees for longitudinal data.
\newblock \emph{Machine Learning}, 106\penalty0 (2):\penalty0 277--305, 2017.

\bibitem[Pedregosa et~al.(2011)Pedregosa, Varoquaux, Gramfort, Michel, Thirion,
  Grisel, Blondel, Prettenhofer, Weiss, Dubourg, Vanderplas, Passos,
  Cournapeau, Brucher, Perrot, and Duchesnay]{scikit}
F.~Pedregosa, G.~Varoquaux, A.~Gramfort, V.~Michel, B.~Thirion, O.~Grisel,
  M.~Blondel, P.~Prettenhofer, R.~Weiss, V.~Dubourg, J.~Vanderplas, A.~Passos,
  D.~Cournapeau, M.~Brucher, M.~Perrot, and {\'E}.~Duchesnay.
\newblock Scikit-learn: Machine learning in python.
\newblock \emph{Journal of Machine Learning Research}, 12\penalty0
  (Oct):\penalty0 2825--2830, 2011.

\bibitem[Quinlan(1986)]{dtinduction}
J.~R. Quinlan.
\newblock Induction of decision trees.
\newblock \emph{Machine Learning}, 1\penalty0 (1):\penalty0 81--106, 1986.

\bibitem[Ramsay and Silverman(2005)]{book_fda}
J.~O. Ramsay and B.~W. Silverman.
\newblock \emph{Functional Data Analysis}.
\newblock Springer, New York, NY, USA, 2005.

\bibitem[Rossi and Villa(2006)]{fda_svm}
F.~Rossi and N.~Villa.
\newblock Support vector machine for functional data classification.
\newblock \emph{Neurocomputing}, 69\penalty0 (1):\penalty0 730--742, 2006.

\bibitem[Sangalli et~al.(2010)Sangalli, Secchi, Vantini, and Vitelli]{fda_kma}
L.~M. Sangalli, P.~Secchi, S.~Vantini, and V.~Vitelli.
\newblock k-mean alignment for curve clustering.
\newblock \emph{Computational Statistics \& Data Analysis}, 54\penalty0
  (5):\penalty0 1219--1233, 2010.

\bibitem[Segal(1992)]{multivrespo}
M.~R. Segal.
\newblock Tree-structured methods for longitudinal data.
\newblock \emph{Journal of the American Statistical Association}, 87\penalty0
  (418):\penalty0 407--418, 1992.

\bibitem[Shang and Hyndman(2018)]{package_fds}
H.~L. Shang and R.~J. Hyndman.
\newblock fds: Functional data sets.
\newblock \url{https://CRAN.R-project.org/package=fds}, 2018.
\newblock R package version 1.8.

\bibitem[Slawski and Hein(2013)]{nnlstsq}
M.~Slawski and M.~Hein.
\newblock Non-negative least squares for high-dimensional linear models:
  consistency and sparse recovery without regularization.
\newblock \emph{Electronic Journal of Statistics}, 7\penalty0 (1):\penalty0
  3004--3056, 2013.

\bibitem[Smyth et~al.(1995)Smyth, Gray, and Fayyad]{kdeleafs}
P.~Smyth, A.~Gray, and U.~M. Fayyad.
\newblock Retrofitting decision tree classifiers using kernel density
  estimation.
\newblock In \emph{International Conference on Machine Learning}, 1995.

\bibitem[Strobl et~al.(2007)Strobl, Boulesteix, and
  Augustin]{ctreemissingvalues}
C.~Strobl, A.-L. Boulesteix, and T.~Augustin.
\newblock Unbiased split selection for classification trees based on the gini
  index.
\newblock \emph{Computational Statistics \& Data Analysis}, 52\penalty0
  (1):\penalty0 483--501, 2007.

\bibitem[Tibshirani and Hastie(2007)]{margintrees}
R.~Tibshirani and T.~Hastie.
\newblock Margin trees for high-dimensional classification.
\newblock \emph{Journal of Machine Learning Research}, 8\penalty0
  (Mar):\penalty0 637--652, 2007.

\bibitem[Torgo(1997)]{regreleafs}
L.~Torgo.
\newblock Functional models for regression tree leaves.
\newblock In \emph{International Conference on Machine Learning}, 1997.

\bibitem[Vens et~al.(2008)Vens, Struyf, Schietgat, D\v{z}eroski, and
  Blockeel]{dthierarchical}
C.~Vens, J.~Struyf, L.~Schietgat, S.~D\v{z}eroski, and H.~Blockeel.
\newblock Decision trees for hierarchical multi-label classification.
\newblock \emph{Machine Learning}, 73\penalty0 (2):\penalty0 185--214, 2008.

\bibitem[W{\"a}chter and Biegler(2006)]{ipopt}
A.~W{\"a}chter and L.~T. Biegler.
\newblock On the implementation of an interior-point filter line-search
  algorithm for large-scale nonlinear programming.
\newblock \emph{Mathematical Programming}, 106\penalty0 (1):\penalty0 25--37,
  2006.

\bibitem[Yamada et~al.(2003)Yamada, Suzuki, Yokoi, and
  Takabayashi]{dttimesstandard}
Y.~Yamada, E.~Suzuki, H.~Yokoi, and K.~Takabayashi.
\newblock Decision-tree induction from time-series data based on a
  standard-example split test.
\newblock In \emph{International Conference on Machine Learning}, 2003.

\bibitem[Yao et~al.(2005)Yao, M{\"u}ller, and Wang]{fda_sparselongdata}
F.~Yao, H.~M{\"u}ller, and J.~Wang.
\newblock Functional data analysis for sparse longitudinal data.
\newblock \emph{Journal of the American Statistical Association}, 100\penalty0
  (470):\penalty0 577--590, 2005.

\bibitem[Yu and Lambert(1999)]{fda_treeresponse_timeofday}
Y.~Yu and D.~Lambert.
\newblock Fitting trees to functional data, with an application to time-of-day
  patterns.
\newblock \emph{Journal of Computational and Graphical Statistics}, 8\penalty0
  (4):\penalty0 749--762, 1999.

\bibitem[Yuan and Cai(2010)]{fda_rkhs}
M.~Yuan and T.~T. Cai.
\newblock A reproducing kernel {H}ilbert space approach to functional linear
  regression.
\newblock \emph{The Annals of Statistics}, 38\penalty0 (6):\penalty0
  3412--3444, 2010.

\bibitem[Zhang(1998)]{ctreemultibinres}
H.~Zhang.
\newblock Classification trees for multiple binary responses.
\newblock \emph{Journal of the American Statistical Association}, 93\penalty0
  (441):\penalty0 180--193, 1998.

\end{thebibliography}

\end{document}